%% file: capturability_notes.tex
\documentclass[journal]{IEEEtran}
\usepackage{import}

\usepackage[cmex10]{amsmath}%
\usepackage{amsfonts}%
\usepackage{amssymb}%
\usepackage{graphicx}
\usepackage{epstopdf}

\usepackage{caption}
\usepackage{subcaption}

\usepackage{algorithmicx}
\usepackage{algpseudocode}

\usepackage{tikz}
\usetikzlibrary{shapes.geometric, arrows}

\usepackage{cite}

\newtheorem{theorem}{Theorem}

\newtheorem{corollary}{Corollary}

\newtheorem{definition}{Definition}

\numberwithin{equation}{section}

\ifCLASSINFOpdf
\else
\fi
  \usepackage[caption=false,font=normalsize,labelfont=sf,textfont=sf]{subfig}
\begin{document}
%
\title{Capturability-based Analysis of Legged Robot with Consideration of Swing Legs}
%
%
%

\author{Zhiwei~Zhang,~\IEEEmembership{Member,~IEEE,}
and Chenglong~Fu,~\IEEEmembership{Member,~IEEE,}

\thanks{Zhiwei Zhang is with the Department of Mechanical Engineering, Tsinghua University, Beijing, China.}
\thanks{Chenglong Fu is with the Department of Mechanical Engineering, Tsinghua University, Beijing, China.}
}

%
%

\markboth{}%
{Zhang \MakeLowercase{\textit{et al.}}: Capturability}
%



\maketitle

\begin{abstract}
Capturability of a robot determines whether it is able to capture a robot within a number of steps. Current capturability analysis is based on stance leg dynamics, without taking adequate consideration on swing leg. 
In this paper, we combine capturability-based analysis with swing leg dynamics. 
We first associate original definition of capturability with a time-margin, which encodes a time sequence that can capture the robot. This time-margin capturability requires consideration of swing leg, and we therefore introduce a swing leg kernel that acts as a bridge between step time and step length. We analyze $N$-step capturability with a combined model of swing leg kernels and a linear inverted pendulum model. 
 By analyzing swing leg kernels with different parameters, we find that more powerful actuation and longer normalized step length result in greater capturability. We also answer the question whether more steps would give greater capturability. For a given disturbance, we find a step sequence that minimizes actuation. This step sequence is whether a step time sequence or a step length sequence, and this classification is based on boundary value problem analysis. 
\end{abstract}


\input{introduction}

\input{capturability_framework}


\input{3D_LIPM_with_Swing_Leg_Kinematics}

\input{Power_Function_Optimization}

\input{algorithm}

\input{discussion}

%
\IEEEpeerreviewmaketitle

\input{appendix}


%



\ifCLASSOPTIONcaptionsoff
  \newpage
\fi

\bibliographystyle{IEEEtran}
\bibliography{mybibtex}
\end{document}

%% file: introduction.tex
\section{Introduction}
Preventing falls is a primary issue in legged locomotion. Not only may it damage a legged robot, a fall will harm its surroundings as well. Even though many current legged robots are able to walk without falling, they have insufficient ability to resist unexpected disturbance. They tend to fall when they slip, when they are pushed or when they lose balance. In comparison, human beings are able to handle most disturbance in everyday life. The ability of legged robots to resist disturbance and prevent falling should be greatly improved to make them competent for universal tasks. 

Stability theory in traditional control theory concerns this ability to resist disturbance. Common technologies are pole placement, phase and gain margin, PID controller. However, these traditional approaches fail to work for a legged system, as a legged system is generally characterized by nonlinear dynamics, under-actuation and a combination of continuous and discrete dynamics. These characterizations result in reduced application of traditional control theories in legged systems. Even the famous Lyapunov stability analysis fails, since legged locomotion does not always exhibit a fixed equilibrium point or a limit cycle. These years, people developed some novel control theories regarding stability in legged locomotion. Some noteworthy theories are zero moment point (ZMP), Poincar\'{e} map analysis and capturability-based analysis. 

ZMP \cite{vukobratovic1969contribution}\cite{vukobratovic2004zero} is the point on the ground where the net ground reaction force produces zero moment in the horizontal direction. If ZMP locates inside support polygon,  stance foot will not turn over. ZMP can be computed theoretically from known joint trajectories or experimentally from force sensors at feet. A common ZMP control strategy is to maintain the ZMP by tracking precomputed reference joint trajectories. ZMP is widely used and largely expanded in walking robots. However, ZMP is not a necessary condition for stable walking. Looking at human walking and running, it is not hard to find that ankle rotates. ZMP is a conservative method, requiring fully contact between stance foot and ground. This limits its application in dynamic walking. 

Walking is mostly periodic, and Poincar\'{e} return map is a powerful tool in periodic stability analysis. Its key idea is to convert a periodic stability problem to a equilibrium point stability problem. Hurmuzlu et al. first used Poincar\'{e} return map in analysis of biped stability\cite{htirmiizlii1987bipedal}. Some follow-on works were reported in passive dynamic walking\cite{mcgeer1990passive}\cite{goswami1996limit}, partial dynamic walking\cite{tedrake2004actuating} and 3D biped walking\cite{grizzle2001asymptotically}. A major challenge in application of Poincar\'{e} map is huge computational cost. Grizzle et al. reduced computation by introducing virtual constrains and hybrid zero dynamics\cite{grizzle2001asymptotically}. However, if actuation fails to maintain the virtual constraints, it may still fall. Poincar\'{e} map approach also fails for non-periodic walking problems. 

Capturability-based analysis is a foot placement estimator. It concerns existence of a control that is able to capture the robot in $N$ steps. Koolen et al.\cite{koolen2012capturability} applied capturability-based analysis on a linear inverted pendulum model, and developed the notion of capture regions. They designed controllers based on this for M2V2 robot\cite{pratt2012capturability}. Zaytsev et al.\cite{zaytsev2015two} concerned capturability for a 2D inverted pendulum with massless legs, and concluded that two step plan is enough. 

As capturability-based analysis discusses where a robot should step, constraints should be considered that describe the robot's ability to swing its foot. From this perspective, the dynamics of swing leg should be taken into account. However, neither \cite{koolen2012capturability} nor \cite{zaytsev2015two} made adequate assumptions on swing leg. \cite{koolen2012capturability} assumed stepping to maximum step length at earliest step time, which is usually contradictory. If it was not successful in doing that, capture region would change. \cite{zaytsev2015two} limited step size and push-off impulse. However, it still provided insufficient information on swing leg dynamics. 
Moreover, these two have difficulties in answering the question what time intervals should be between steps. The intervals between steps are critical in step planning, in that the state of the robot will possibly involve to an unstable state if it takes too much time to finish a step.  

Our analysis in this paper is based on a combined model of a linear inverted pendulum model and a model describing swing leg dynamics. We expand definition of capturability to a notion of capturability that is assigned with a time sequence. We call it time-margin capturability. This notion of capturability ensures that the robot is capturable under this specified step time sequence. We continue by introducing a swing leg kernel that describes kinematics of swing leg. We view swing leg kernel simply as a power function of swing time, and is proportional to actuation coefficient, which describes overall influence from swing leg actuation. We then combine swing leg kernel with a linear inverted pendulum model for capturability analysis.  We analyze how coefficient of actuation and maximum step length influence capturability, and also plan stepping when a disturbance is given. We finish with comparison of our results with those of others.

%% file: capturability_framework.tex
\section{Capturability Framework}
\label{Capturability Framework}

We begin with definition of capturability, and then explain why we need to add a time-margin with it.
\begin{definition}[captured state]
A \emph{captured state} is the state in which the robot stops but does not fall.
\label{captured state}
\end{definition}

\begin{definition}[capturable state]
A state of the robot $X(t)$ is said to be \emph{capturable} if there is a feasible control $u(T)$, for $T \geq t$, that drives the robot to a captured state without ever reaching $X_{fall}$.
\label{def: capturable state}
\end{definition}

\begin{definition}[N-step capturable state]
A state $X(t)$ is said to be \emph{N-step capturable} if the control as described in Definition~\ref{def: capturable state} is one with at least $N$ step location-time pairs $\left(r_i, t_i \right)$, where $i \geq N$.
\label{def: N-step capturable state}
\end{definition}

We remind readers that our definition of $N$ step capturability is different from that defined in \cite{koolen2012capturability}. 
Definition~\ref{def: N-step capturable state} uses \emph{at least} to define capturability, thus making a  $N$-step capturable state incapturable with less than $N$ steps. We find our definition easier to be understood, by saying, for example, a 2-step capturable state cannot be captured with a single step, but is possible with 3 steps.

\begin{corollary}[iterative description of N-step capturability]
A step is able to  drive a $N$-step capturable state to a $(N-1)$-step capturable state. 
\label{corollary:iterative def}
\end{corollary}

From Corollary \ref{corollary:iterative def}, a step is able to drive a $N$-step capturable state to a $(N-1)$-step capturable state. However, this doesn't necessitate that the new state will definitely be $(N-1)$-step capturable after a step; it depends on its state and how well the step is. This motivates the following definition of $N$-step capture point (and then capture region) which describes how well the step is. Capture region also serves to move capturability framework from state space to $2$-D Euclidean space. 

\begin{definition}[$N$-step capture point at time $t$]
We talk about the notion of capture point for a state $X(t)$ at time $t$ in Euclidean space $\mathbb{R}^2$. Suppose at time $t$, the robot is in a $M$-step capturable state $X(t)$. $N$ is a interger such that $M\leq N$. A point $x \in \mathbb{R}^2$ is an \emph{N-step capture point} for the robot associated with state $X(t)$ at time $t$, if the action of stepping to $x$ at $t$ drives the robot to a $(N-1)$-step capturable state $X(t^+)$ of the same time $t$.
\label{def: N-step capturable point}
\end{definition}

The condition that $M \leq N$ is a result of `at least' from Definition \ref{def: N-step capturable state}. An $M$-step capturable state can be captured within more than $M$ steps, but not less. Thus, an $M$-step capturable state is associate with $N$-capture points where $N\geq M$. We rephase this in the following corollary.

\begin{corollary}
Suppose a state $X(t)$ is $M$-step capturable. There exist $N$-step capture points in $\mathbb{R}^2$, where $N \geq M$.
\end{corollary} 
This corollary relates capturability in terms of state space to capturability in terms of Euclidean space. 

Definition~\ref{def: N-step capturable point} gives a time-vaying view of a point $x\in \mathbb{R}^2$ being $N$-step capturable or not. It ensures that the new state is $(N-1)$-step capturable at the time of step, $i.e.\ $at time $t^+$. However, it doesn't guarantee that the robot is always $(N-1)$-step capturable in the future. Without further prompt and proper actions, it is possible that the state $X(t^+)$ evolves to a state $X(t+\delta)$ which is not $(N-1)$-step capturable. How large $\delta$ could be can be considered as a metric of how well the step is. Hinted by this, we define a version of $N$-step capturability framework with a time margin associated to it.

We begin with 1-step time-margin capturability.
\begin{definition}[$\Delta_1$-margin $1$-step capturable state]
Suppose at time $t_0$, the state is $X(t_0)$ and a time period $\Delta_1$ is associated with the robot. The state $X(t_0)$ is said to be \emph{$\Delta_1$-margin $1$-step capturable} if there is a step, taken at time $t_1=t_0+\Delta_1$, drives the robot to a captured state.
\label{def:margin 1-step}
\end{definition}

How large this $\Delta_1$ can be is a property of $X(t_0)$ and the ability of the robot to swing its foot. 

With this, we give an iterative definition of $N$-step capturable state with time margins.
\begin{definition}[$(\Delta_{N} \cdots \Delta_1)$-margin $N$-step capturable]
Suppose at time $t_{N}$, $N \geq1$, the state is $X(t_{N})$ and a sequence of time periods $\Delta_1, \cdots, \Delta_{N}$ is associated. The robot is said to be \emph{$\Delta_{N} \cdots \Delta_1$-margin $N$-step capturable} if there is a step, taken at $t_{N-1}=t_{N}+\Delta_N$, drives the robot to a $(\Delta_{N-1} \cdots \Delta_1)$-margin $(N-1)$-step capturable state and remains till the following step at time $t_{N_2}=t_{N-1}+\Delta_{N-1}$. The location of the step $r_{n-1} \in \mathbb{R}^2$ is called a \emph{$\Delta_{N} \cdots \Delta_1$-margin $N$-step capture point} of state $X(t_{N})$.
\label{def:iterative N-step}
\end{definition}

\section{3D LIPM Model with Point Foot}
\label{point foot}
In this section, we focus on 3D LIPM Model with point foot. 
Its dynamic model is proposed in \cite{koolen2012capturability}. We include their dimensionless results here for our convenience. 
\begin{eqnarray}
\ddot{r}_{CoM} &=& \mathbf{P}r_{CoM} - r_{ankle}
\label{eq:dyn 1} \\
r_{ic} &=& \mathbf{P}r_{CoM} + \dot{r}_{CoM}
\label{eq: dyn 2} \\
\dot{r}_{ic} &=& r_{ic} - r_{ankle}
\label{eq: dyn 3}
\end{eqnarray}
where $r_{CoM}$ is normalized center of mass; $r_{ic}$ is normalized location of instantaneous capture center; $r_{ankle}$ is normalized ankle location. See \cite{koolen2012capturability} for details in normalization.

Consider LIPM after a step. Denote the time right after a step as $t_{s0}$. $r_{ic}(t_{s0})$ can be calculated from Equation~\eqref{eq: dyn 2}. With this initial condition, trajectory of instantaneous capture point is 
\begin{equation}
r_{ic}(t) = \left[ r_{ic}(t_{s0}) - r_{ankle} \right] \mathbf{e}^t + r_{ankle}
\label{eq:ic evolve}
\end{equation}
By now we have a way to find the instantaneous capture point for any state of the model, as a result, we have the following theorem.
\begin{theorem}[existence and uniqueness of instantaneous capture point]
For a LIPM,  the instantaneous capture point exists and is unique.
\label{thm:existence of icp}
\end{theorem}

\begin{theorem}[existence and uniqueness of 1-step capture point]
For LIPM with point foot, the instantaneous capture point is the unique 1-step capture point, if it is able to step to any location at any time.
\label{thm:existence of 1-step capture point}
\end{theorem}
\begin{IEEEproof}
See Appendix \ref{proof uniqueness}.
\end{IEEEproof}

\begin{corollary}
For a LIPM, any state is 1-step capturable, if it is able to step to any location at any time.
\label{clr:LIPM 1-step capturable}
\end{corollary}

Theorem~\ref{thm:existence of 1-step capture point} and Corollary~\ref{clr:LIPM 1-step capturable} assumes that a LIPM can step to any location at any time. This is a significantly aggressive and usually inappropriate assumption. In real scenarios, limitations, such as maximum step length and step speed, should be considered, and  Corollary~\ref{clr:LIPM 1-step capturable} needs modifications. When talking about $N$-step capturability, we find it is more useful and meaningful to use definition of time-margin $N$-step capturability. Unlike definition of $N$-step capturability as in \cite{koolen2012capturability}, time-margin $N$-step capturability provides more information. Not only does it answer the question whether it is $N$-step capturable, it also specifies a step sequence that captures the robot in $N$ steps. Thus, each step is connected under its corresponding time-margin. 

This time-margin is influenced by the ability of the robot to swing foot. LIPM model considers only stance leg, containing no information on stepping ability. Thus, in order to adopt this time-margin notion of capturability, we have to delve into the swing leg that performs stepping.


%% file: 3D_LIPM_with_Swing_Leg_Kinematics.tex
\section{3D LIPM with Swing Leg Kinematics}
\label{slk}
\subsection{swing leg kernal}
It is intuitive to assume that how far the robot is able to step is in relation to how long it takes to perform the step. Under same actuation, it should take more time to step further. Thus, we assume that dimensionless\footnote{We use dimensionless analysis in this section. All terms are normalized as those in Section~\ref{point foot}.} step length $l$ 
is a continuous and increasing function of dimensionless step time interval $\tau$\footnote{It can be easily shown that this $\tau$ is the time-margin between two consecutive steps.}, denoted as 
\begin{equation}
l = f_k(\tau)
\label{eq:assumption}
\end{equation}
and we term it \emph{swing leg kernal}.
To make this assumption more realistic, we further limit $\tau$. $\tau$ is upper bounded by $\tau_{max}=f_k^{-1}(l_{max})$, where $l_{max}$ is maximum normalized step length of the robot.

\subsection{1-step capturability with swing leg kernel}
We begin with derivation regarding 1-step capturability. We separate $x$ and $y$ axes, as they are decoupled in LIPM model. This degrades our problem from 2D to 1D, and greatly simplifies the problem.

\begin{figure}%
\centering
\def\svgwidth{0.9\columnwidth}
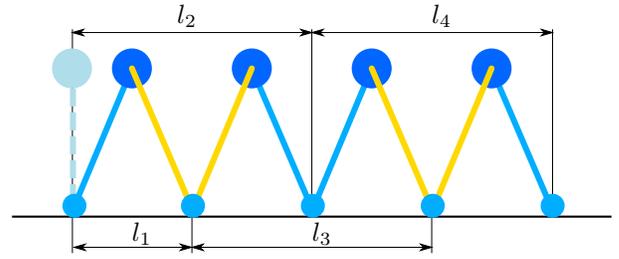
\caption{step pattern for a biped robot. 
Low saturation blue indicates initial position $x=0$ of the robot. A push happens at $t=0$, and the robot starts to step at $t=0^+$. Two legs are dyed blue and yellow respectively. $l_1$ to $l_4$ represent each step length in a chronological order, where $l_i=f_k(\tau_i)$.}%
\label{fig:walking}%
\end{figure}

Suppose that the robot stands at $x=0$ (see Fig.~\ref{fig:walking}). At time $t=0$, a push disturbance acts on the robot, resulting the instantaneous capture point at $x_{ic}(0) = d$.  Then we have by \eqref{eq:ic evolve}
\begin{equation}
x_{ic} (\tau)= x_{ic}(0) \mathbf{e}^{\tau}
\label{eq:1-step slk}
\end{equation}where $\tau \in \left[ 0, \tau_{max} \right]$.

For 1-step case, left hand side of \eqref{eq:1-step slk} is step length for 1-step, $i.e.$ the $l$ in \eqref{eq:assumption}, and $x_{ic}(0)$ is $d$. This gives
\begin{equation}
d = \frac{f_k(\tau)}{\mathbf{e}^{\tau}}
\label{eq:d=frac}
\end{equation}

From \eqref{eq:d=frac}, we see that $d$ is a continuous function of $\tau$, and $\tau$ is closed, thus $d$ has a maximum $d_{max}$ at $\tau_{opt}$. If the initial $d > d_{max}$, there is no real solution of \eqref{eq:d=frac}, and the related state is not 1-step capturable. 
\begin{theorem}[1-step capturable with swing leg kernel]
A state is \emph{1-step capturable with swing leg kernel} described by \eqref{eq:assumption} if and only if \eqref{eq:d=frac} has a solution $\tau\in \mathbb{R}$. This $\tau$ is the related time-margin and is generally not unique. Existence of $\tau$ is also equivalent to $d \leq d_{max}$. Step length $l$ defined by \eqref{eq:assumption} with $\tau$ is a 1-step capture point.
\label{thm:1-step slk}
\end{theorem}

It is noteworthy that \eqref{eq:d=frac} can be treated as a comparison between swing leg kinematics and stance leg kinematics. If swing leg moves faster, then numerator is greater with same $\tau$, thus $d$ is greater. It is a kind of competition between swing leg and stance leg kinematics. From this fact, we hold that, when talking about capturability, it is of advantage for a legged robot to have faster swing legs.

\subsection{$N$-step capturability with swing leg kernel}
\label{$N$-step slk}
Suppose we use two steps to capture a robot, each step takes a time interval of $\tau_1$ and $\tau_2$. For the first step, we have
\begin{equation}
x_{ic}(\tau_1) = d \mathbf{e}^{\tau_1}
\label{eq:}
\end{equation}
and for the second step, we have from \eqref{eq:ic evolve}
\begin{equation}
x_{ic}(\tau_1 + \tau_2) = \left[ x_{ic}(\tau_1) - f_k(\tau_1)\right] \mathbf{e}^{\tau_2} + f_k(\tau_1)
\label{eq:4.4}
\end{equation}
If the robot is captured with these two steps, it should satisfy
\begin{equation}
x_{ic}(\tau_1 + \tau_2) = f_k(\tau_2)
\label{eq:4.6}
\end{equation}
By equating \eqref{eq:4.4}\eqref{eq:4.6}, we have
\begin{equation}
d=\frac{f_k(\tau_1)}{\mathbf{e}^{\tau_1}} + \frac{-f_k(\tau_1)+f_k(\tau_2)}{\mathbf{e}^{(\tau_1+\tau_2)}} 
\label{eq:}
\end{equation}
Similarly, in a 3-step capturable case, for the third step, we have
\begin{equation}
x_{ic}(\tau_1 + \tau_2 + \tau_3) = \left[ x_{ic}(\tau_1+\tau_2) - f_k(\tau_2)\right] \mathbf{e}^{\tau_3} + f_k(\tau_2)
\label{eq:4.7}
\end{equation}
If the robot is captured with these three steps, it should satisfy
\begin{equation}
x_{ic}(\tau_1 + \tau_2 + \tau_3) = f_k(\tau_1) + f_k(\tau_3)
\label{eq:4.8}
\end{equation}
By equating \eqref{eq:4.7}\eqref{eq:4.8}, we have
\begin{equation}
d=\frac{f_k(\tau_1)}{\mathbf{e}^{\tau_1}} + \frac{-f_k(\tau_1)+f_k(\tau_2)}{\mathbf{e}^{(\tau_1+\tau_2)}} + \frac{f_k(\tau_1)-f_k(\tau_2)+f_k(\tau_3)}{\mathbf{e}^{(\tau_1+\tau_2+\tau_3)}}
\label{eq:}
\end{equation}
Iteratively, we have a result for $N$-steps,
\begin{equation}
d_N = \sum\limits_{i=1}^N{\sum\limits_{j=1}^i{\left(-1\right)^{i+j} f_k(\tau_j)}   \mathbf{e}^{ -\sum\limits_{j=1}^i{\tau_j}}}
\label{eq:n-step slk}
\end{equation}
By now we have similiar result as that with Theorem~\ref{thm:1-step slk}. For a $N$-step capturable state, we can find a \emph{sequence of step time} and consequently a \emph{sequence of step length}.
\begin{corollary}[$N$-step capturable with swing leg kernel]
A state is \emph{$N$-step capturable with swing leg kernel} discribed by \eqref{eq:assumption} if and only if \eqref{eq:n-step slk} has a solution $\tau_i\in \mathbb{R}, i=1,2, \cdots, N$. $\tau_i$ are the time-margins. $l_i$  defined by \eqref{eq:assumption} with $\tau_i$ is a sequence of $N$-step capture points. Existence of $\tau_i$ is also equivalent to $d_N \leq \max \limits_{\tau_i}d_N$. 
\label{thm:N-step slk}
\end{corollary}
\subsection{optimal step sequence}
For a $N$-step capturable state, solution of \eqref{eq:n-step slk} is generally not unique. To find a unique optimal step sequence, we implement an optimization-based approach to find the optimal step time (and equivalently length) sequence. We define an objective function $f_{obj}$. Now, we are able to formulate our problem. Given an initial instantaneous capture point position $d$ and swing leg kernel as a function of swing time $l=f(\tau)$, find a sequence of time-margin $\tau_i$, $i=1, \cdots, N$, that
\begin{equation}
\min_{\tau_i} f_{obj}
\label{general optimization problem}
\end{equation}
\begin{equation}
\textrm{s.t.} \quad \sum\limits_{i=1}^N{\sum\limits_{j=1}^i{\left(-1\right)^{i+j} f_k(\tau_j)}   \mathbf{e}^{ -\sum\limits_{j=1}^i{\tau_j}}} = d
\end{equation}
\begin{equation}
\tau_i \in \left[ 0, \tau_{max} \right]
\label{eq:}
\end{equation}

%% file: drawing3.eps_tex
\begingroup%
  \makeatletter%
  \providecommand\color[2][]{%
    \errmessage{(Inkscape) Color is used for the text in Inkscape, but the package 'color.sty' is not loaded}%
    \renewcommand\color[2][]{}%
  }%
  \providecommand\transparent[1]{%
    \errmessage{(Inkscape) Transparency is used (non-zero) for the text in Inkscape, but the package 'transparent.sty' is not loaded}%
    \renewcommand\transparent[1]{}%
  }%
  \providecommand\rotatebox[2]{#2}%
  \ifx\svgwidth\undefined%
    \setlength{\unitlength}{1200bp}%
    \ifx\svgscale\undefined%
      \relax%
    \else%
      \setlength{\unitlength}{\unitlength * \real{\svgscale}}%
    \fi%
  \else%
    \setlength{\unitlength}{\svgwidth}%
  \fi%
  \global\let\svgwidth\undefined%
  \global\let\svgscale\undefined%
  \makeatother%
  \begin{picture}(1,0.40903496)%
    \put(0,0){\includegraphics[width=\unitlength]{drawing3.eps}}%
    \put(0.2,0.022){\color[rgb]{0,0,0}\makebox(0,0)[lb]{\smash{$l_1$}}}%
    \put(0.5,0.022){\color[rgb]{0,0,0}\makebox(0,0)[lb]{\smash{$l_3$}}}%
    \put(0.275,0.385){\color[rgb]{0,0,0}\makebox(0,0)[lb]{\smash{$l_2$}}}%
    \put(0.7,0.385){\color[rgb]{0,0,0}\makebox(0,0)[lb]{\smash{$l_4$}}}%
  \end{picture}%
\endgroup%

%% file: Power_Function_Optimization.tex
\section{Swing Leg Kernel as Power Functions}
In this section, we substantiate our discussion in Section~\ref{slk}. We use a family of power functions to describe swing leg kernels, and use an optimization-based method to analyze how the maximum disturbance that the robot can resist changes with respect to swing leg kernels with different coefficients.
\subsection{swing leg kernel as a power function of swing time}
We describe swing leg kinematics as a power function of swing time
\begin{equation}
l = k\tau^a
\label{eq:pf}
\end{equation}
where $\tau$ is normalized time, $l$ is normalized step length, $k\in \mathbb{R}$ is the actuation coefficient, and $a \in \mathbb{R}$ is the index number. We find $k$ and $a$ by simulation. $k$ has a positive relationship with swing torque, and $a$ is around 1.66. See Appendix~\ref{appendix:slk} for details.

With this swing leg kernel, and assuming torque coefficient $k$ are constant during steps, \eqref{eq:n-step slk} becomes 
\begin{equation}
d = k \sum\limits_{i=1}^N{\sum\limits_{j=1}^i{\left(-1\right)^{i+j} \tau_i^{a}   \mathbf{e}^{ -\sum\limits_{j=1}^i{\tau_j}}}}
\label{eq:pfslk}
\end{equation}
The assumption of equal $k$ is reasonable, if the robot makes same effort to swing its leg in every step.

\subsection{influence of actuation and maximum step length on capturability}
\label{vary kl}
In this subsection, we analyze how actuation coefficient $k$ and normalized maximum step length $l_{max}$ influence the ability to resist disturbance $d$. This analysis answers the question: how to select actuation and maximum step length that will give a robot more capturability. We discretize $k$ and $l_{max}$ over a reasonable range, and find the maximum disturbance $d_{max}$ for each $k$-$l_{max}$ combination. This $d_{max}$ describes the maximum ability of the robot to resist disturbance. In this case, our problem \eqref{general optimization problem} becomes: for each $k$ and $l_{max}$ combination,
\begin{equation}
\max_{\tau_i} d
\end{equation}
\begin{equation}
\textrm{s.t.} \quad  k \sum\limits_{i=1}^N{\sum\limits_{j=1}^i{\left(-1\right)^{i+j} \tau_i^{a}   \mathbf{e}^{ -\sum\limits_{j=1}^i{\tau_j}}}} = d
\label{eq:pfslk}
\end{equation}
\begin{equation}
k\tau_i^{a} \leq l_{max}
\label{eq:}
\end{equation}
This is the same problem as 
\begin{equation}
\max \quad  \sum\limits_{i=1}^N{\sum\limits_{j=1}^i{\left(-1\right)^{i+j} \tau_i^{a}   \mathbf{e}^{ -\sum\limits_{j=1}^i{\tau_j}}}}
\label{eq:max d in t}
\end{equation}
\begin{equation}
\textrm{s.t.} \quad  \tau_i \leq \sqrt[a]{\frac{l_{max}}{k}}
\label{feasible set t}
\end{equation}

\begin{figure}
    \centering
    \begin{subfigure}[b]{0.24\textwidth}
        \includegraphics[width=\textwidth]{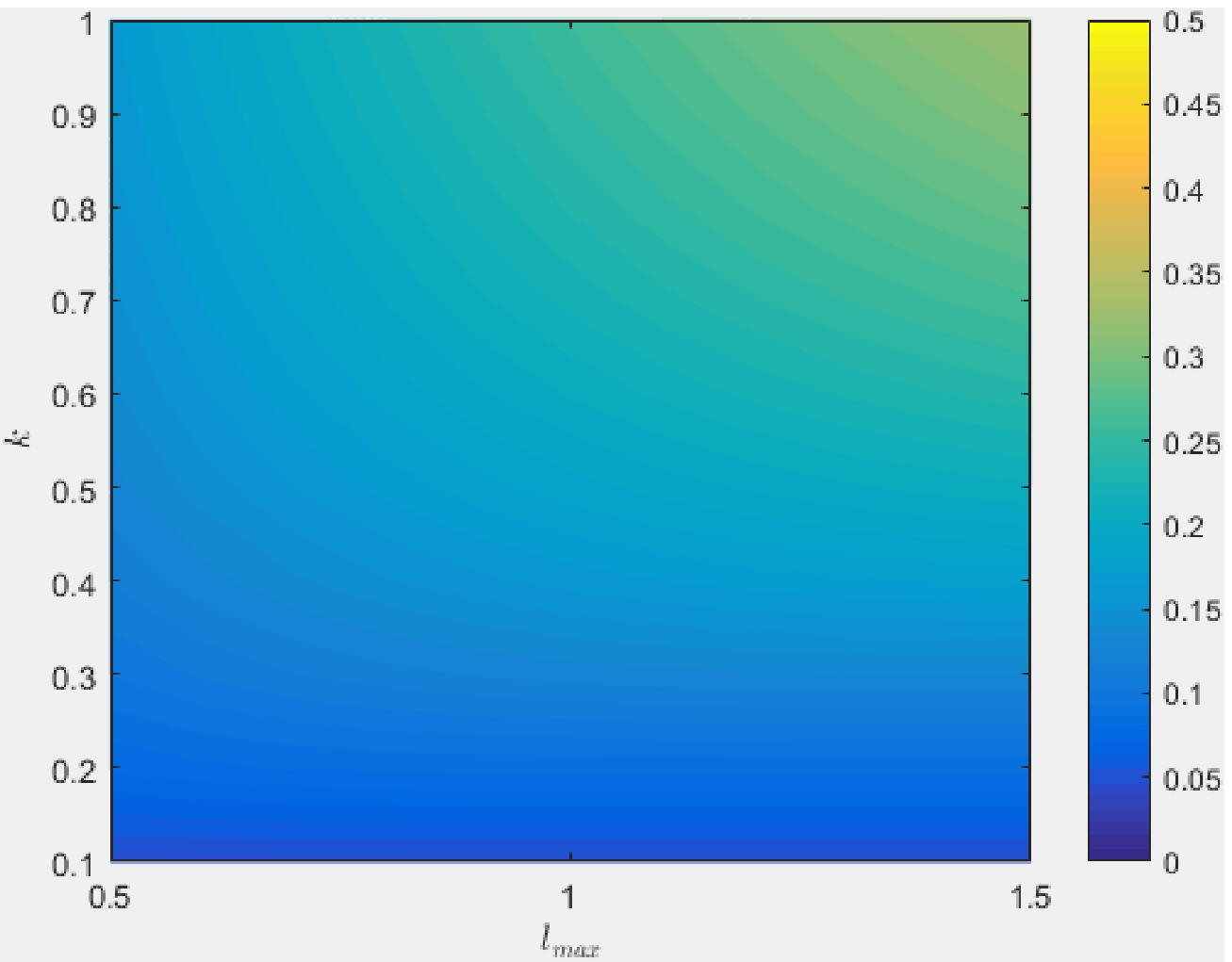}
        \caption{}
        \label{fig:k-l matrix 1}
    \end{subfigure}
    \begin{subfigure}[b]{0.24\textwidth}
        \includegraphics[width=\textwidth]{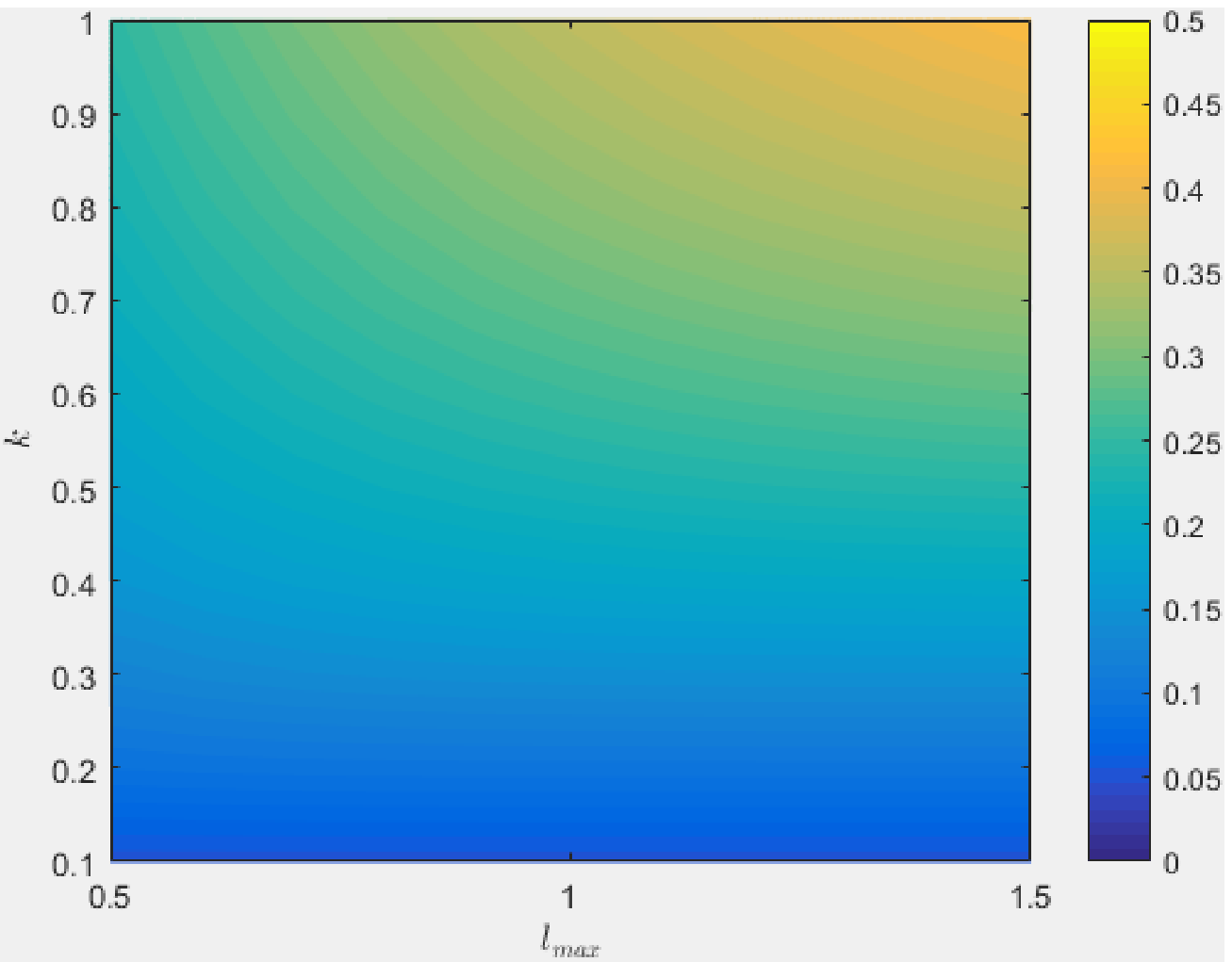}
        \caption{}
        \label{fig:k-l matrix 2}
    \end{subfigure}
    \begin{subfigure}[b]{0.24\textwidth}
        \includegraphics[width=\textwidth]{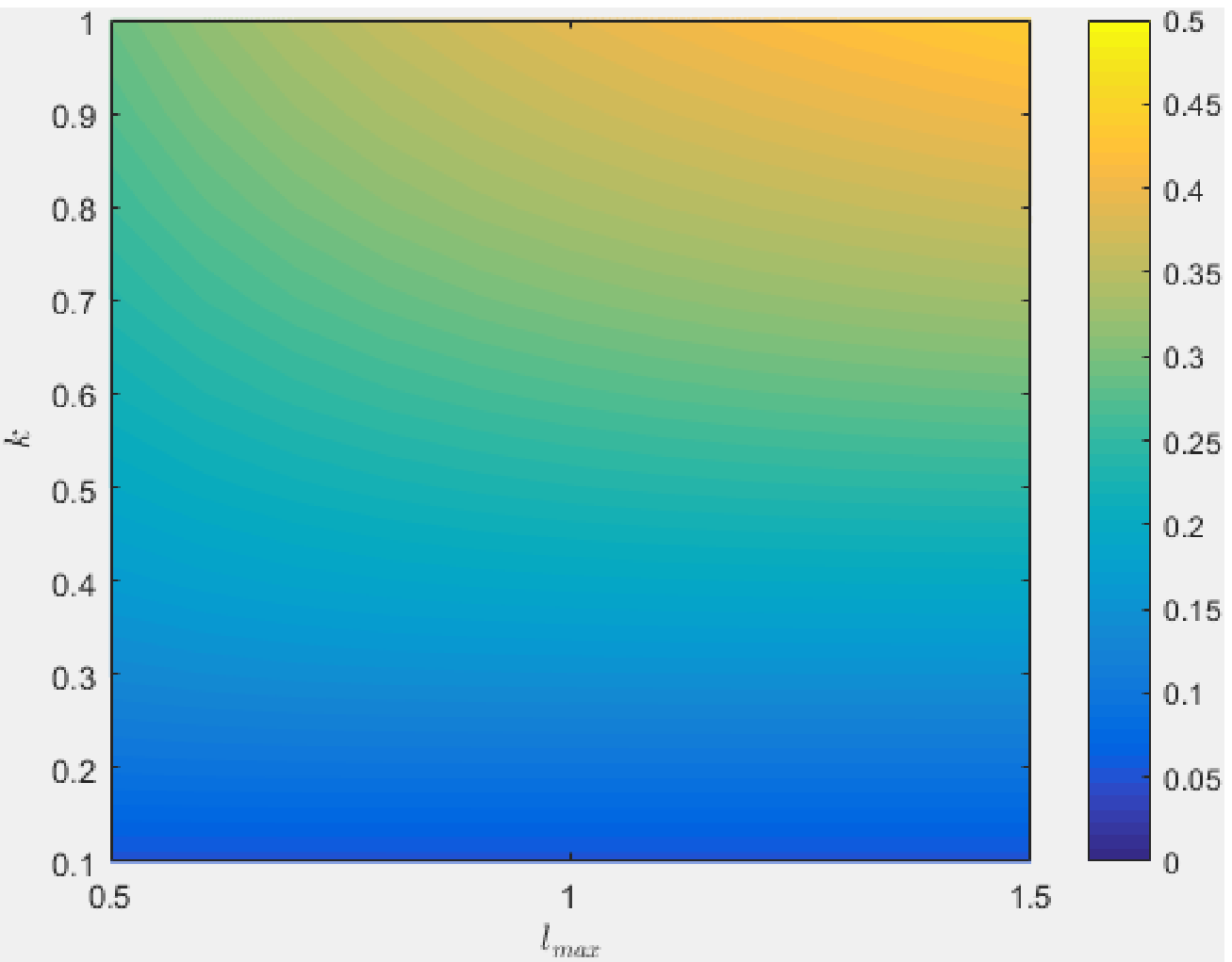}
        \caption{}
        \label{fig:k-l matrix 3}
    \end{subfigure}
		\begin{subfigure}[b]{0.24\textwidth}
        \includegraphics[width=\textwidth]{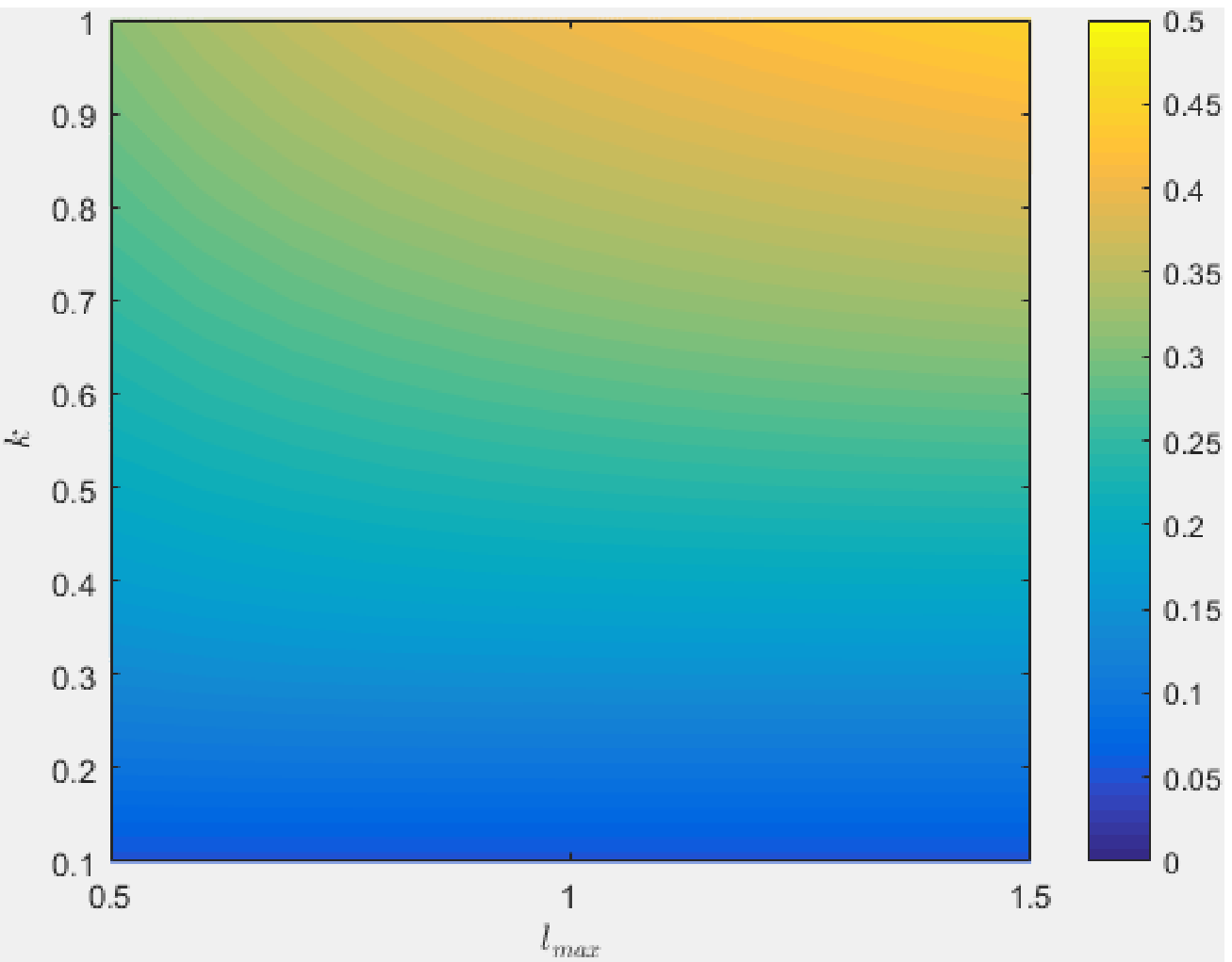}
        \caption{}
        \label{fig:k-l matrix 4}
    \end{subfigure}
    \caption{$d_{max}$ of different $k$ and $l_{max}$ for		(a) 1-step; (b) 2-step; (c) 3-step; (d) 4-step.
		}
		\label{k-l matrix}
\end{figure}

We find the maximum $d$ for each $k$-$l_{max}$ combination up to 4-step case, and visualize results in Fig.~\ref{k-l matrix}. In each subfigure, $x$ axis is $l_{max}$, and $y$ axis is $k$. Value of $d_{max}$ as a function of $k$ and $l_{max}$ is illustrated by different colors. All four subfigures shares same colorbar, with red largest and blue smallest. 

For each subfigure in Fig.~\ref{k-l matrix}, maximum disturbance  that the robot is able to resist $d_{max}$ is non-decreasing with respect to $k$ and $l_{max}$ respectively. Generally speaking, with fixed $l_{max}$, $d_{max}$ increases as $k$ increases; with fixed $k$, $d_{max}$ increases as $l_{max}$ increases. A special case is when $k$ is very small, where $d_{max}$ remains the same as $l_{max}$ increases. With these results, we conclude that it is of benefit to select stronger actuation and greater maximum nominal step length, in order for the robot to resist larger disturbance.

\begin{figure}
    \centering
    \begin{subfigure}[b]{0.24\textwidth}
        \includegraphics[width=\textwidth]{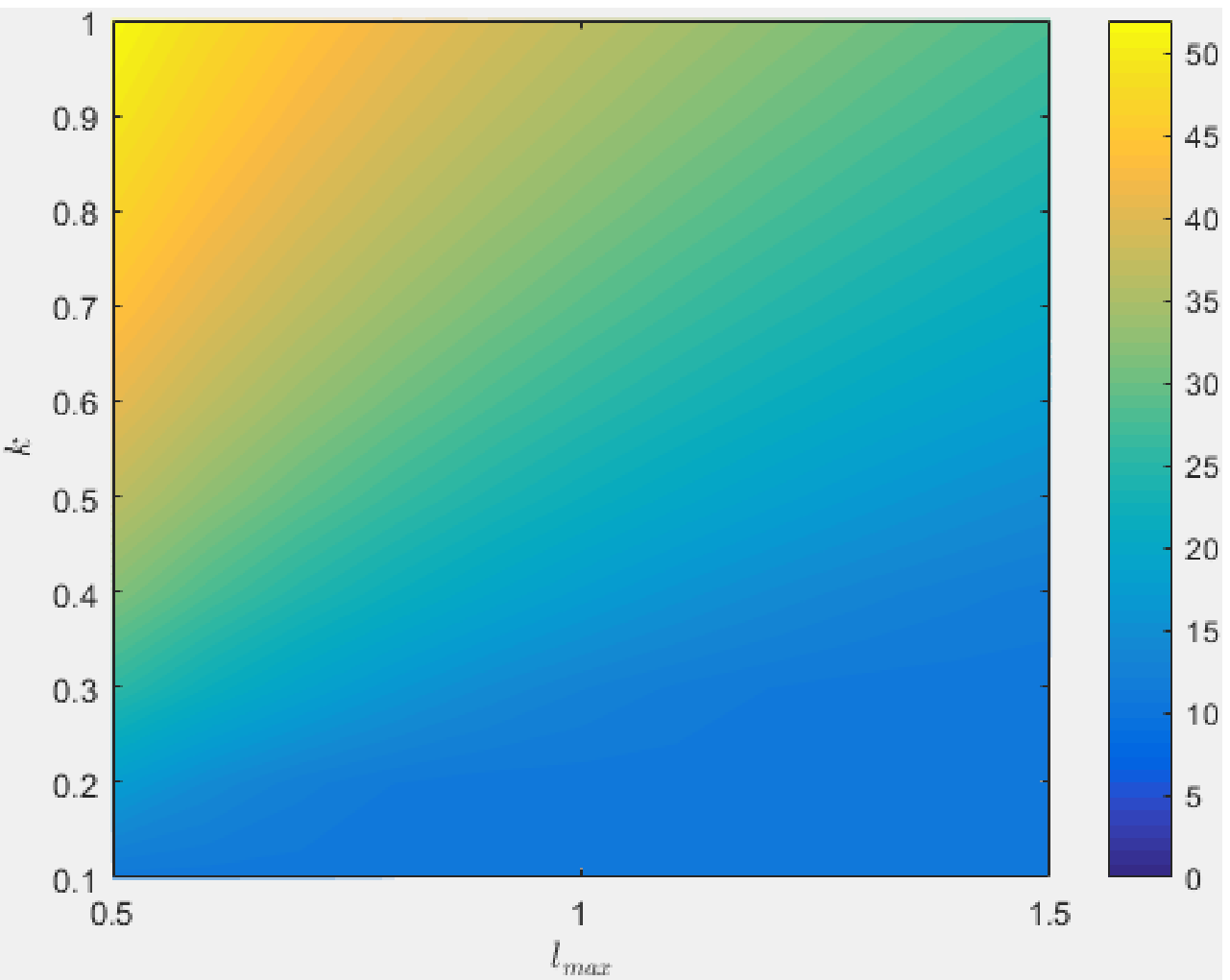}
        \caption{}
        \label{fig:inc k-l matrix 1}
    \end{subfigure}
    \begin{subfigure}[b]{0.24\textwidth}
        \includegraphics[width=\textwidth]{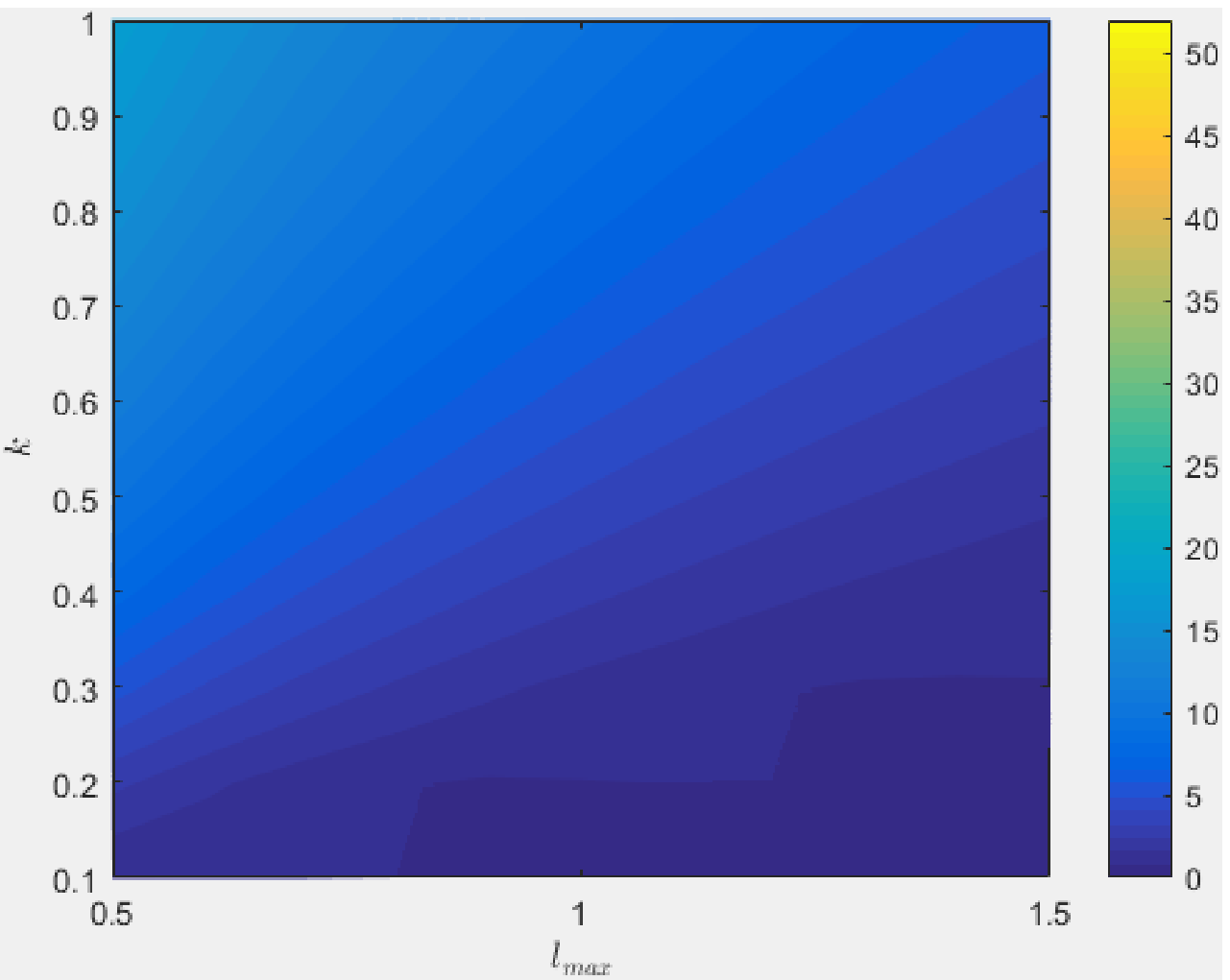}
        \caption{}
        \label{fig:inc k-l matrix 2}
    \end{subfigure}
    \begin{subfigure}[b]{0.27\textwidth}
        \includegraphics[width=\textwidth]{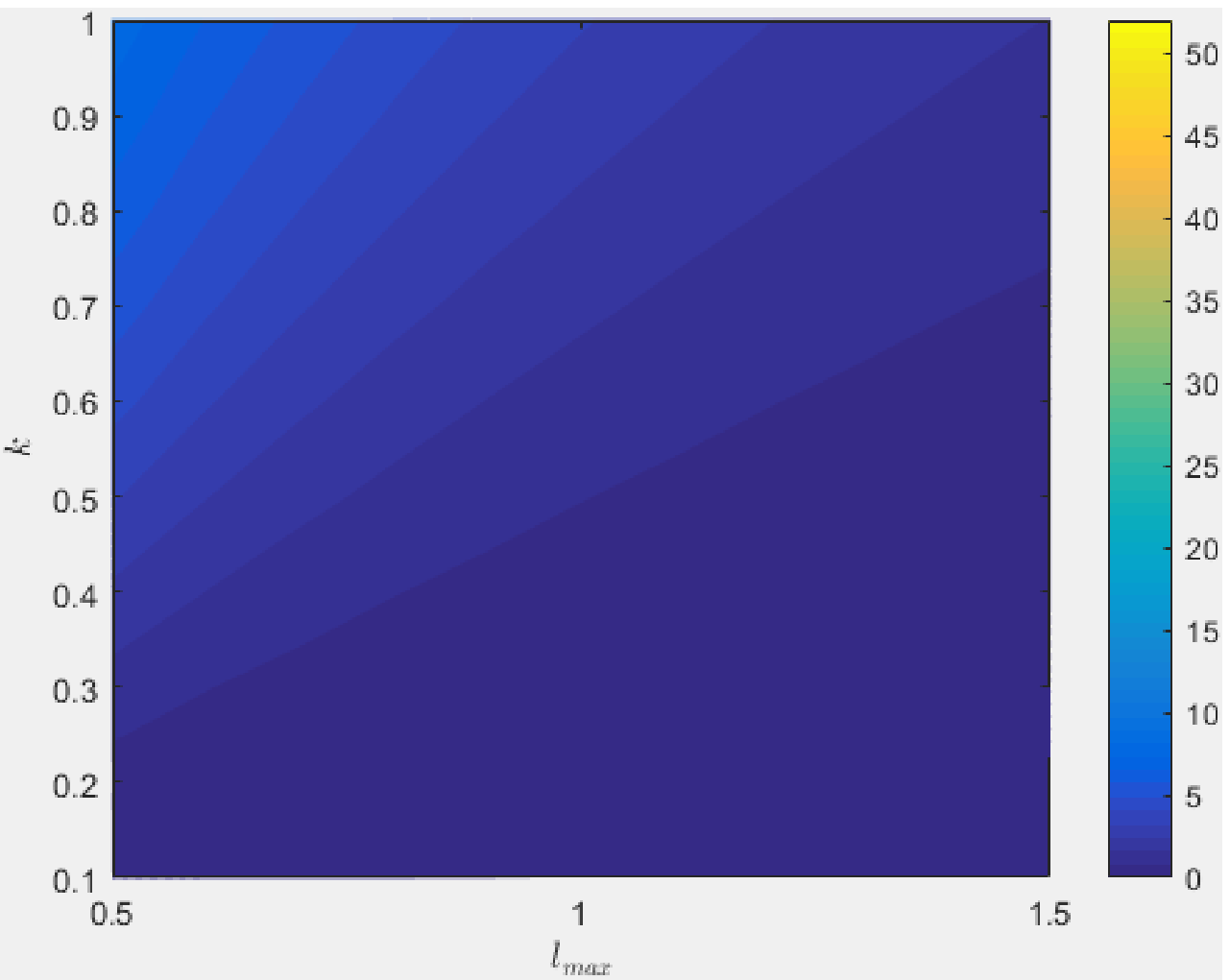}
        \caption{}
        \label{fig:inc k-l matrix 3}
    \end{subfigure}
    \caption{relative incremental percentage in $d_{max}$ between steps 
		(a) from 1-step to 2-step; (b) from 2-step to 3-step; (c) from 3-step to 4-step.
		}
		\label{fig:increase}
\end{figure}

We also compare $d_{max}$ among $N$-step capturability for $N = 1, \cdots, 4$. We summarize our observation in Table~\ref{table:increase} and Fig.~\ref{fig:increase}. For each $k$-$l_{max}$ combination, $d_{max}$ increases from 1-step to 4-step. 
\begin{table}[!t]
\renewcommand{\arraystretch}{1.3}
\caption{relative increase in $d_{max}$ among multiple steps}
\label{table:increase}
\centering
\begin{tabular}{c|c|c|c}
\hline
\bfseries step number & \bfseries minimum  & \bfseries maximum & \bfseries average\\
\hline
from 1 to 2 & 9.84\%& 34.10\% & 19.02\%\\
from 2 to 3 & 0.73\% & 15.00\%& 4.96\%\\
from 3 to 4 & 0.05\% & 7.21\% &  1.58\% \\
\hline
\end{tabular}
\end{table}
We observe significant increase in $d_{max}$ from 1-step to 2-step, with at least 9.84\%. From 2-step to 3-step, this percentage varies from 0.73\% to 15.00\%.  How much relative increase is depends on $k$ and $l_{max}$. We observe large relative increase from 2-step to 3-step for cases where $k$ is large and $l_{max}$ is small. We also notice that even large $k$ with small $l_{max}$ gives greater increase, its absolute value is still small compared to some $k$-$l_{max}$ combinations where $k$ is smaller and $l_{max}$ is larger. Physically speaking, it means that if actuation is  strong but step length is short, it is able to capture a state with possibly over 4 steps. From 3-step to 4-step, mostly we observe less than 5\% increase, and we conclude that a forth step is not necessary for most $k$-$l_{max}$ combinations, except for those with large $k$ and small $l_{max}$. 

For each $k$-$l_{max}$ combination and when taking about $N$-step capturability, we are able to find a step sequence that contains $N$ steps and that maximizes disturbance $d$. In our analysis on its hidden mathematics, we further find that this step sequence has two comings, and we call them whether a step time sequence or a step length sequence. \footnote{The name `step time sequence' and `step length sequence' only represent their mathematical coming. These two sequences are related by swing leg kernel, and both are a sequence of step time and also a sequence of step length.} This step time or step length classification is the result of different feasible set \eqref{feasible set t}, which is influenced by $k/l_{max}$ ratio. For better illustration and explanation, we exemplify with 2-step capturability. 

\begin{figure}
    \centering
		    \begin{subfigure}[b]{0.24\textwidth}
        \includegraphics[width=\textwidth]{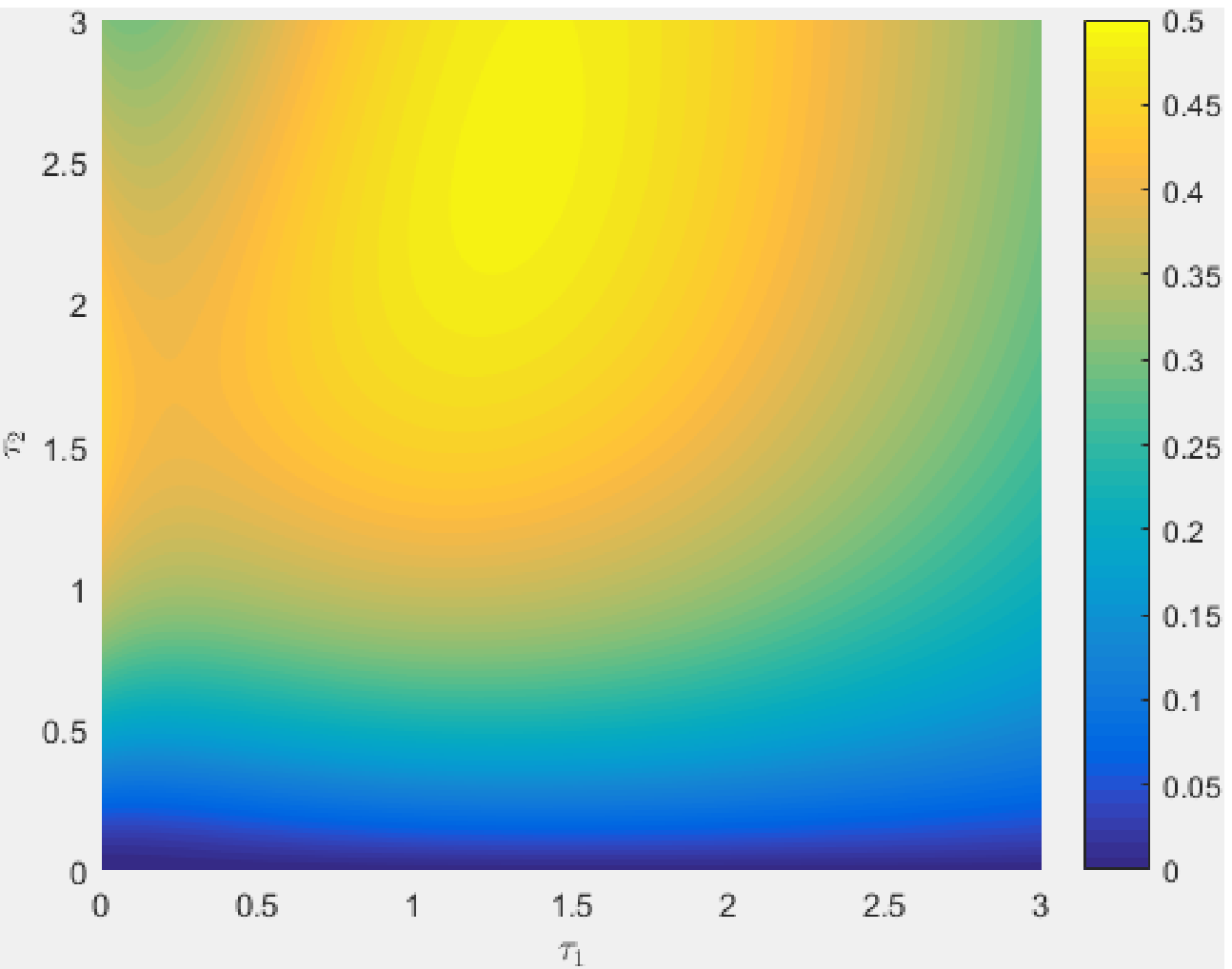}
        \caption{}
        \label{fig:nocons1}
    \end{subfigure}
    \begin{subfigure}[b]{0.24\textwidth}
        \includegraphics[width=\textwidth]{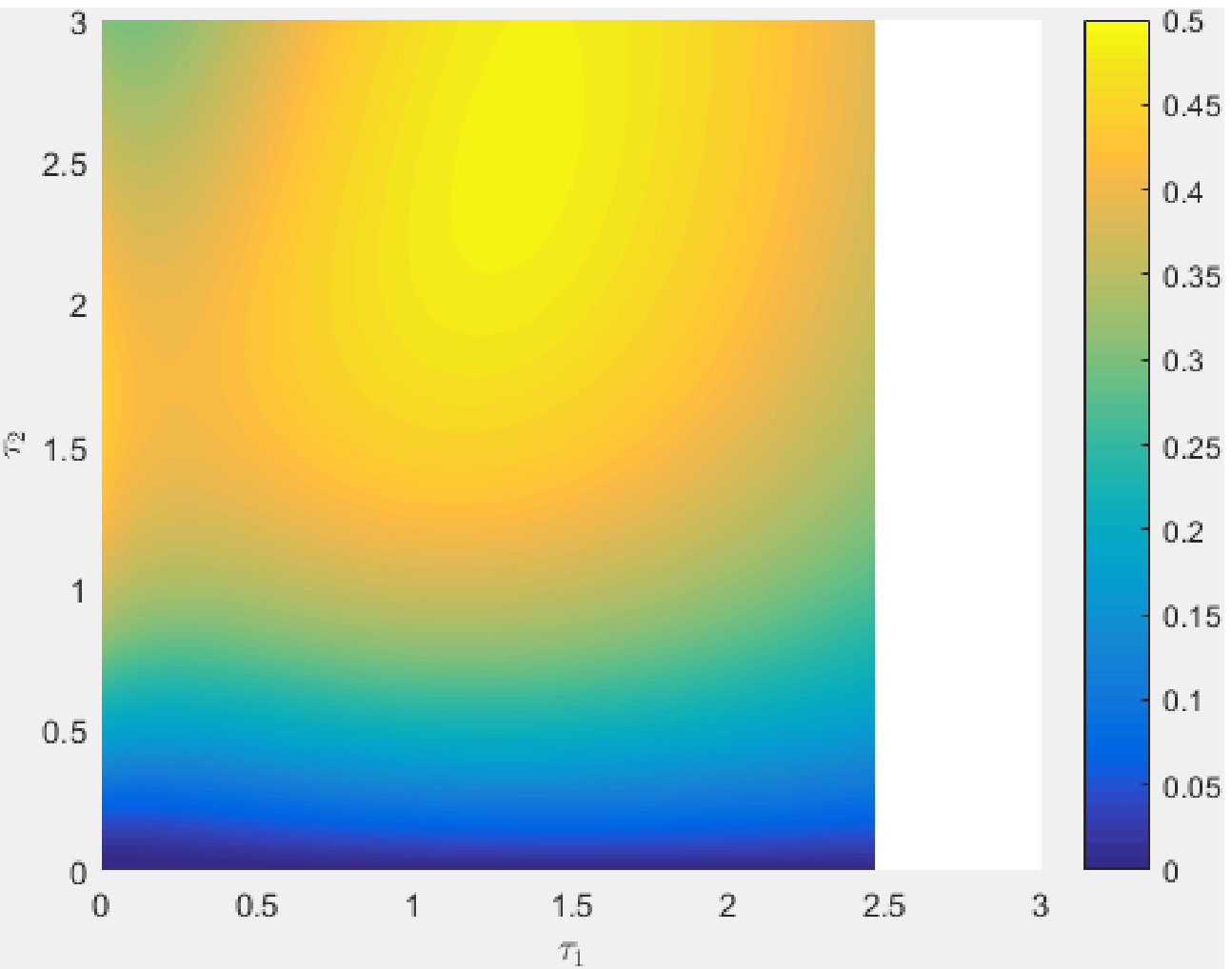}
        \caption{}
        \label{fig:lk9}
    \end{subfigure}
    \begin{subfigure}[b]{0.24\textwidth}
        \includegraphics[width=\textwidth]{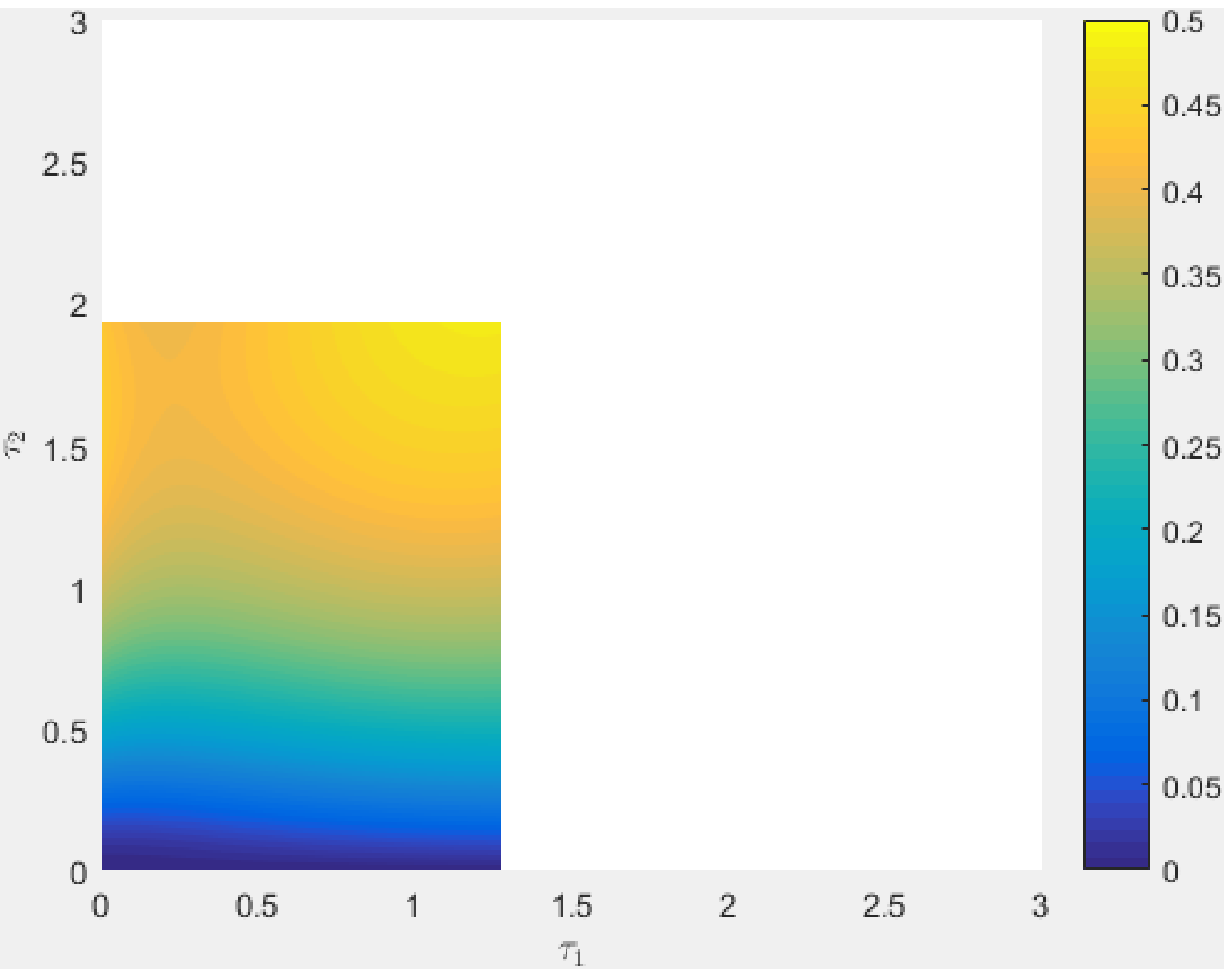}
        \caption{}
        \label{fig:lk3}
    \end{subfigure}
		\begin{subfigure}[b]{0.24\textwidth}
        \includegraphics[width=\textwidth]{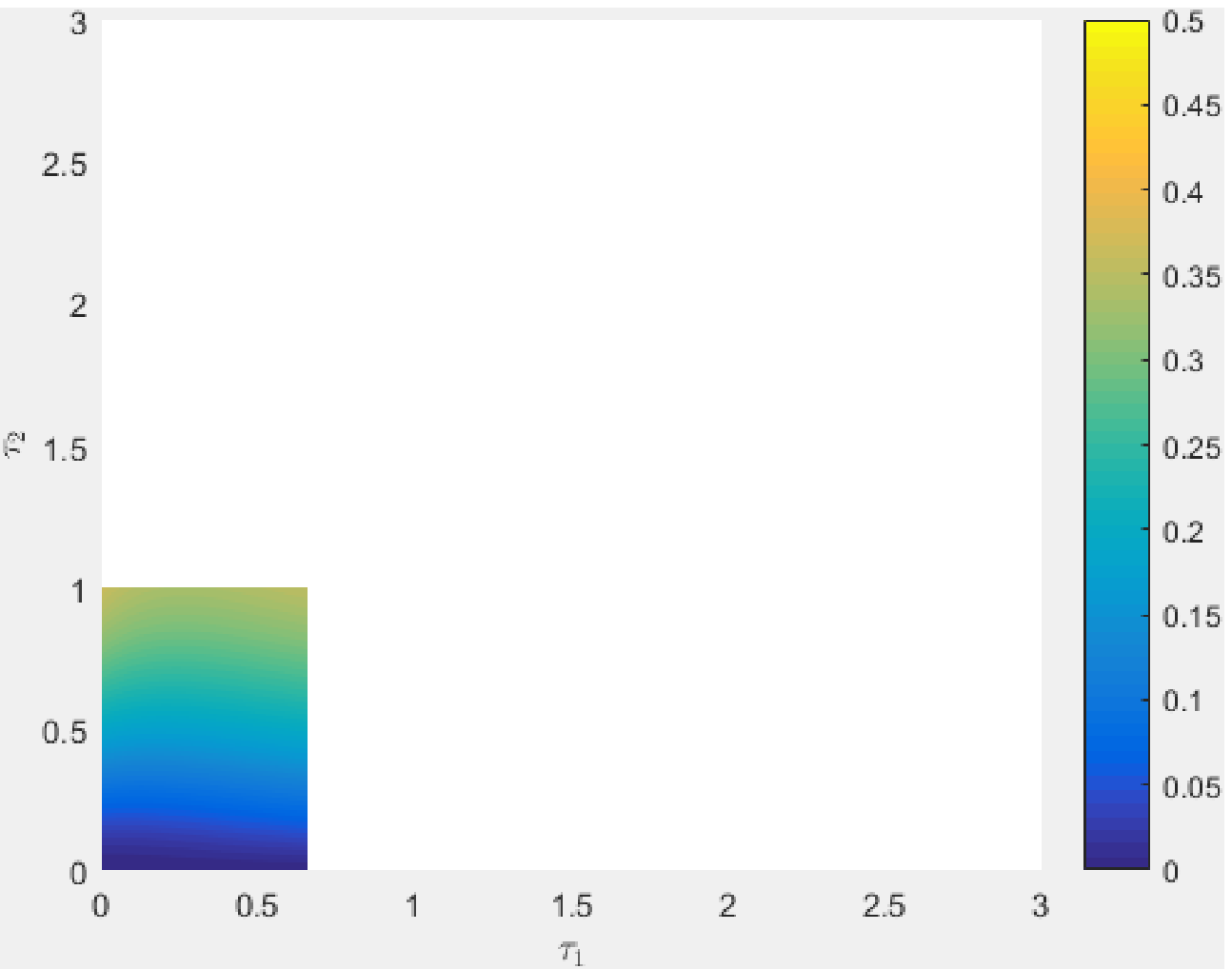}
        \caption{}
        \label{fig:lk1}
    \end{subfigure}
    \caption{influence of $l_{max}/k$ on feasible set, when 
		(a) unconstrained; (b) $l_{max}/k = 9$; (c)  $l_{max}/k = 3$; (d) $l_{max}/k = 1$. 
		}
		\label{feasible set}
\end{figure}

For 2-step capturability, we should look at 2D time domain, with $x$ axis being $\tau_1$, and $y$ axis $\tau_2$. The objective function \eqref{eq:max d in t} is the same for different $k$-$l_{max}$ combinations. Fig.~\ref{fig:nocons} shows objective function value over its domain. For different $l_{max}/k$ ratios, the feasible sets are different. Figure~\ref{fig:lk9} to~\ref{fig:lk1} show how feasible sets looks like for different $l_{max}$ and $k$, and regions that do not satisfy constraints \eqref{feasible set t} are left blank. Different shapes of feasible sets result in whether a step time sequence or a step length sequence. If $l_{max}/k$ is large enough, the global maximum is included in the feasible set, and gives an optimal step time sequence. If $l_{max}/k$ is small enough, a local and also feasible global\footnote{We notice that there is another local optimum at around (0,1.5). In some cases, optimization toolbox may find this local optimum. This local optimum does not satisfy our stepping pattern in Section~\ref{$N$-step slk}, as the first step should be non-zero. We would remove these zero-first-step results, making the optimum at corner the global optimum.} maximum lies at corner, and this gives an optimal step sequence, and each step is $l_{max}$. 

%% file: algorithm.tex
\section{Algorithm Framework for Push Recovery}
Based on our analysis in Section~\ref{slk}, we discuss a preliminary algorithm for push recovery. We find the optimum step sequence by minimizing actuation. As actuation coefficient $k$ increases as actuation increases, our problem becomes: given a disturbance $d$, 
\begin{equation}
\min_{\tau_i} k
\end{equation}
\begin{equation}
\textrm{s.t.} \quad  k \sum\limits_{i=1}^N{\sum\limits_{j=1}^i{\left(-1\right)^{i+j} \tau_i^{a}   \mathbf{e}^{ -\sum\limits_{j=1}^i{\tau_j}}}} = d
\label{eq:pfslk}
\end{equation}
\begin{equation}
 k\tau_i^{a} \leq l_{max}
\label{eq:}
\end{equation}
\begin{equation}
k \in \left[0, k_{max} \right]
\label{eq:kmax}
\end{equation}
where $k_{max}$ describes the maximum actuation the robot is able to output. 

This problem is equivalent to 
\begin{equation}
\max_{\tau_i} \quad f_{obj} =   \sum\limits_{i=1}^N{\sum\limits_{j=1}^i{\left(-1\right)^{i+j} \tau_i^{a}   \mathbf{e}^{ -\sum\limits_{j=1}^i{\tau_j}}}}
\label{eq:pfslk}
\end{equation}
\begin{equation}
\textrm{s.t.} \quad  \frac{\tau_i^{a}}{f_{obj}} \leq \frac{l_{max}}{d}
\label{eq:feasible set kmin}
\end{equation}

and minimum $k_{min}$ is found by 
\begin{equation}
k_{min} = \frac{d}{f_{obj,max}}
\label{eq:}
\end{equation}
If $k_{min} \leq k_{max}$, then the robot is $N$-step capturable; otherwise, it is not $N$-step capturable.

\begin{figure}%
\centering
 \begin{subfigure}[b]{0.24\textwidth}
        \includegraphics[width=\textwidth]{figure/fs_nocons}
        \caption{}
        \label{fig:nocons}
    \end{subfigure}
    \begin{subfigure}[b]{0.24\textwidth}
        \includegraphics[width=\textwidth]{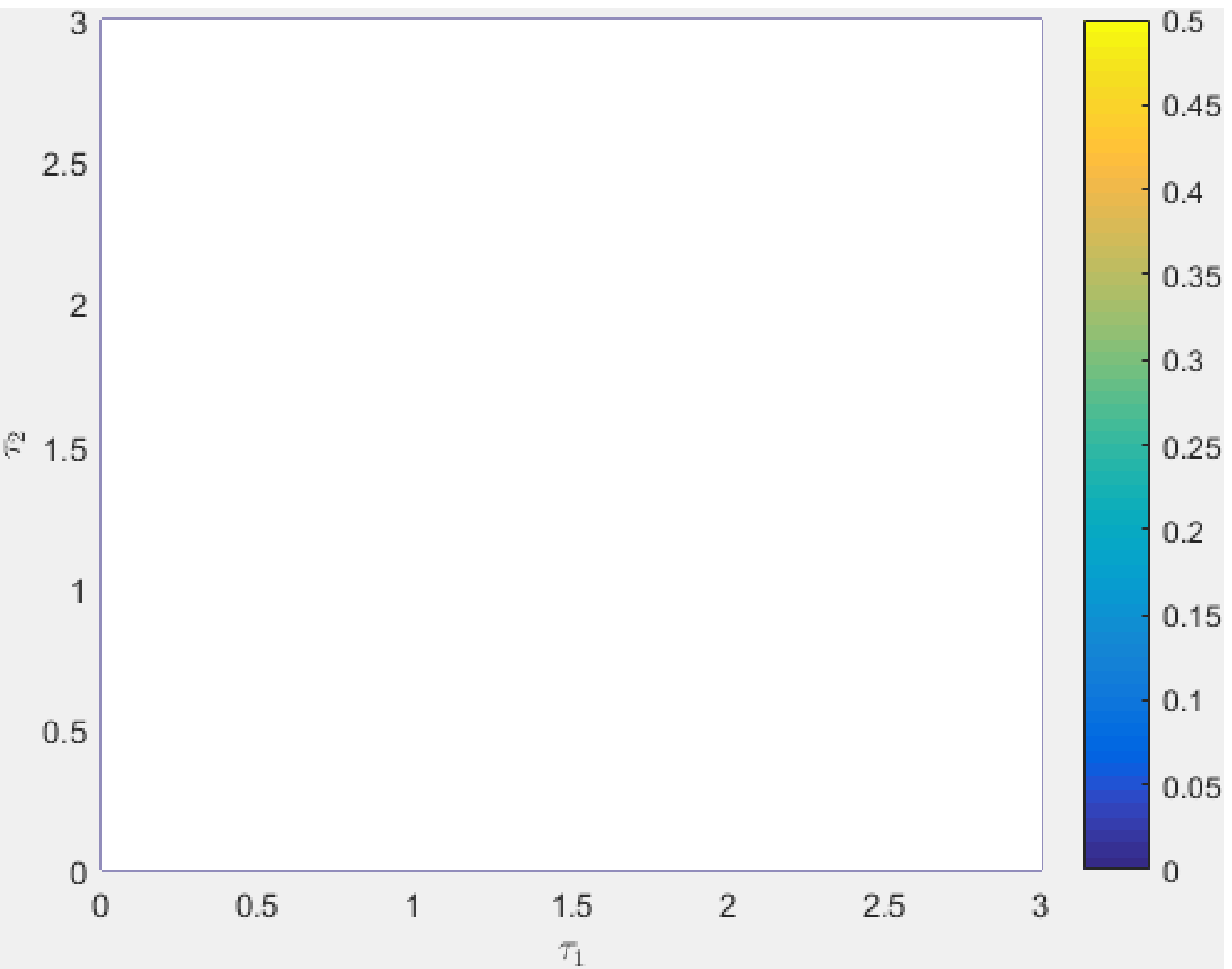}
        \caption{}
        \label{fig:ld1}
    \end{subfigure}
    \begin{subfigure}[b]{0.24\textwidth}
        \includegraphics[width=\textwidth]{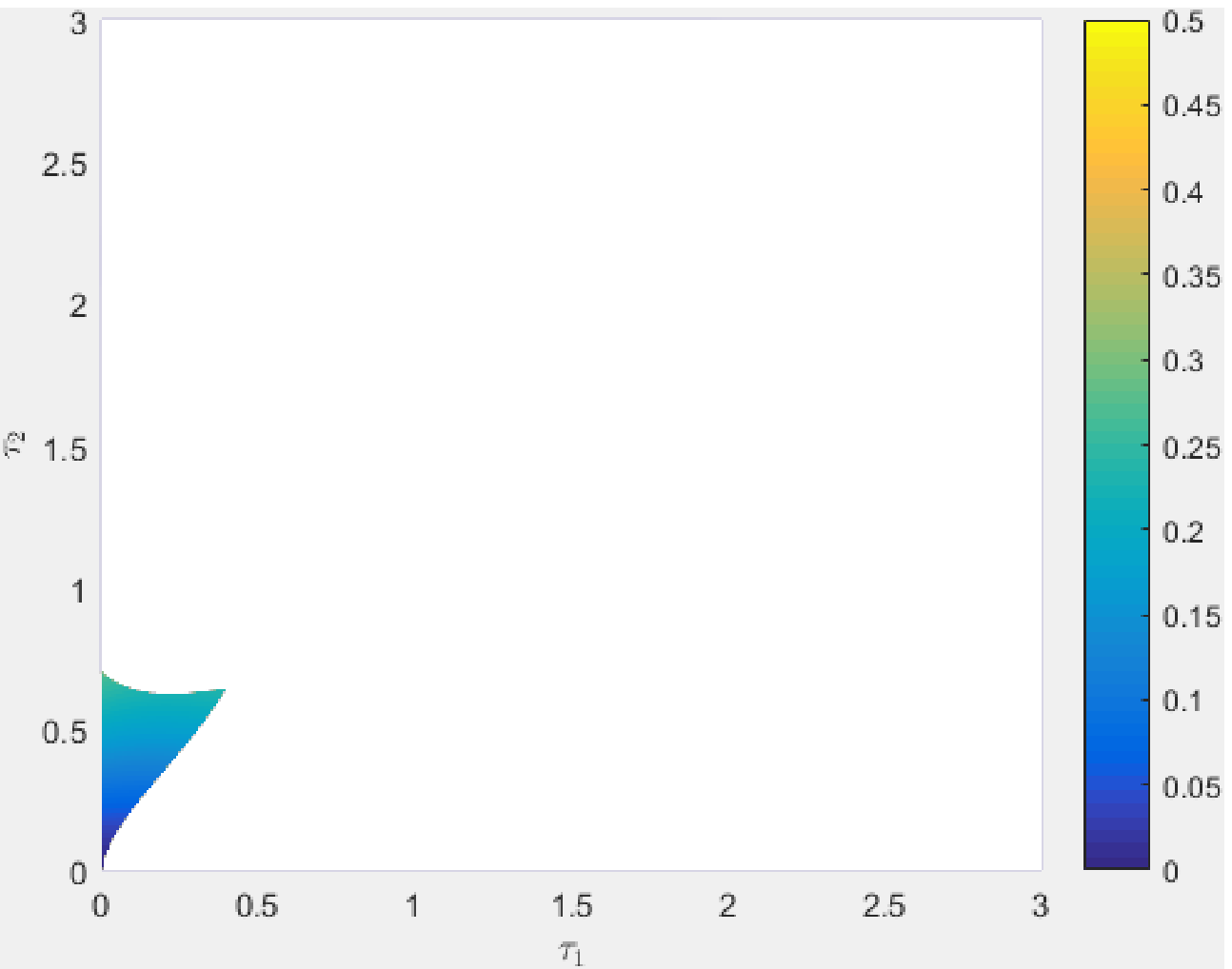}
        \caption{}
        \label{fig:ld2}
    \end{subfigure}
		\begin{subfigure}[b]{0.24\textwidth}
        \includegraphics[width=\textwidth]{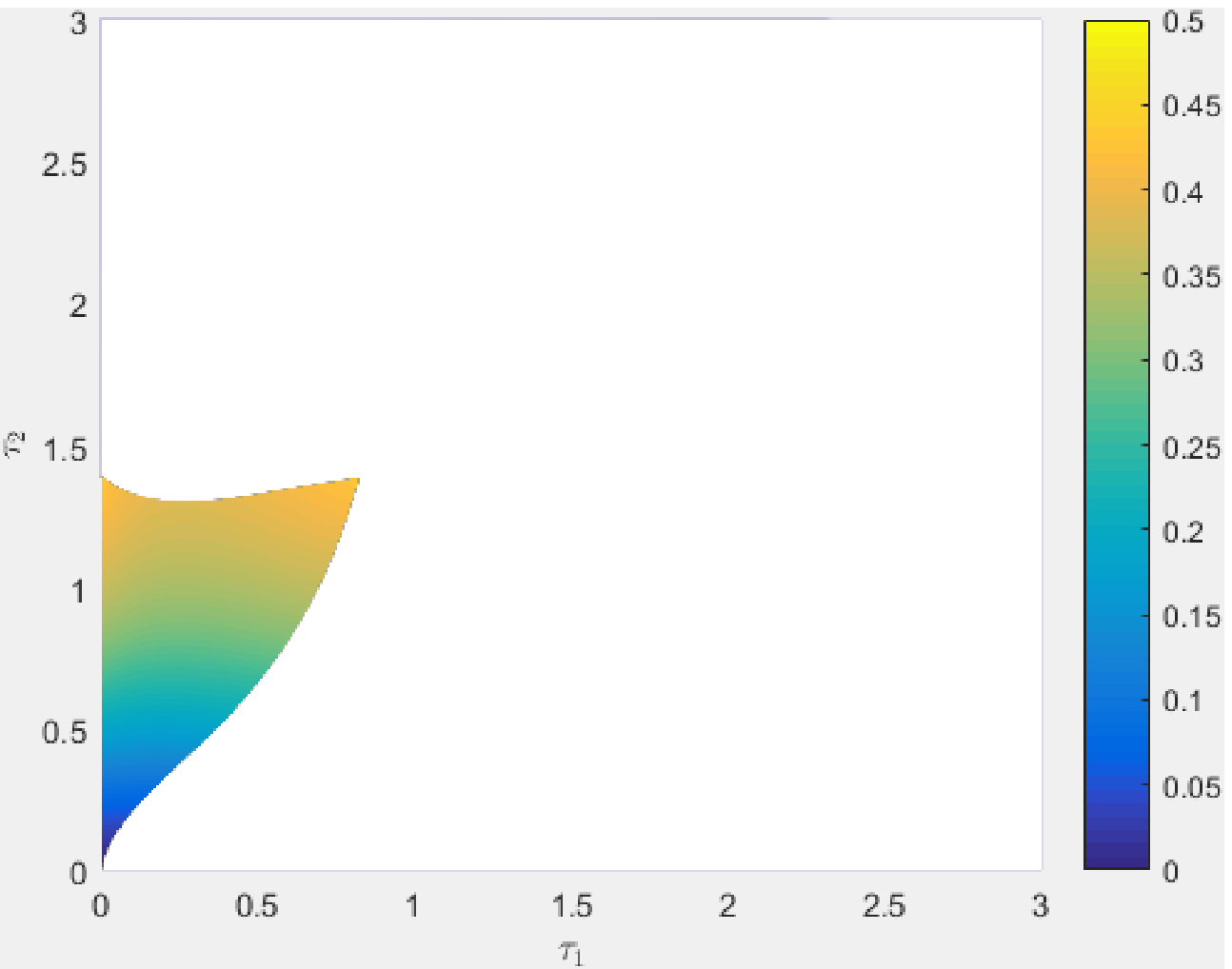}
        \caption{}
        \label{fig:ld4}
    \end{subfigure}
				\begin{subfigure}[b]{0.24\textwidth}
        \includegraphics[width=\textwidth]{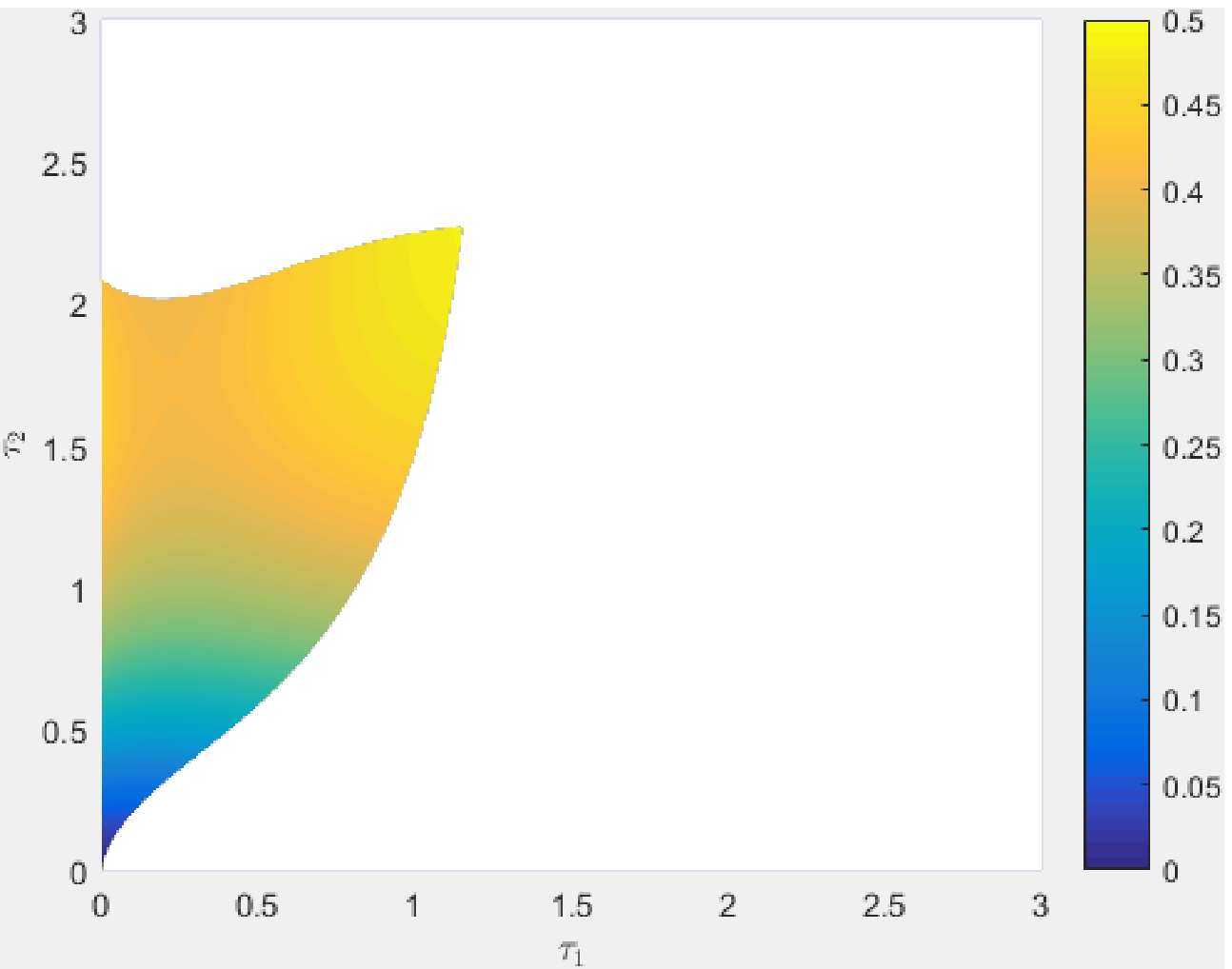}
        \caption{}
        \label{fig:ld8}
    \end{subfigure}
				\begin{subfigure}[b]{0.24\textwidth}
        \includegraphics[width=\textwidth]{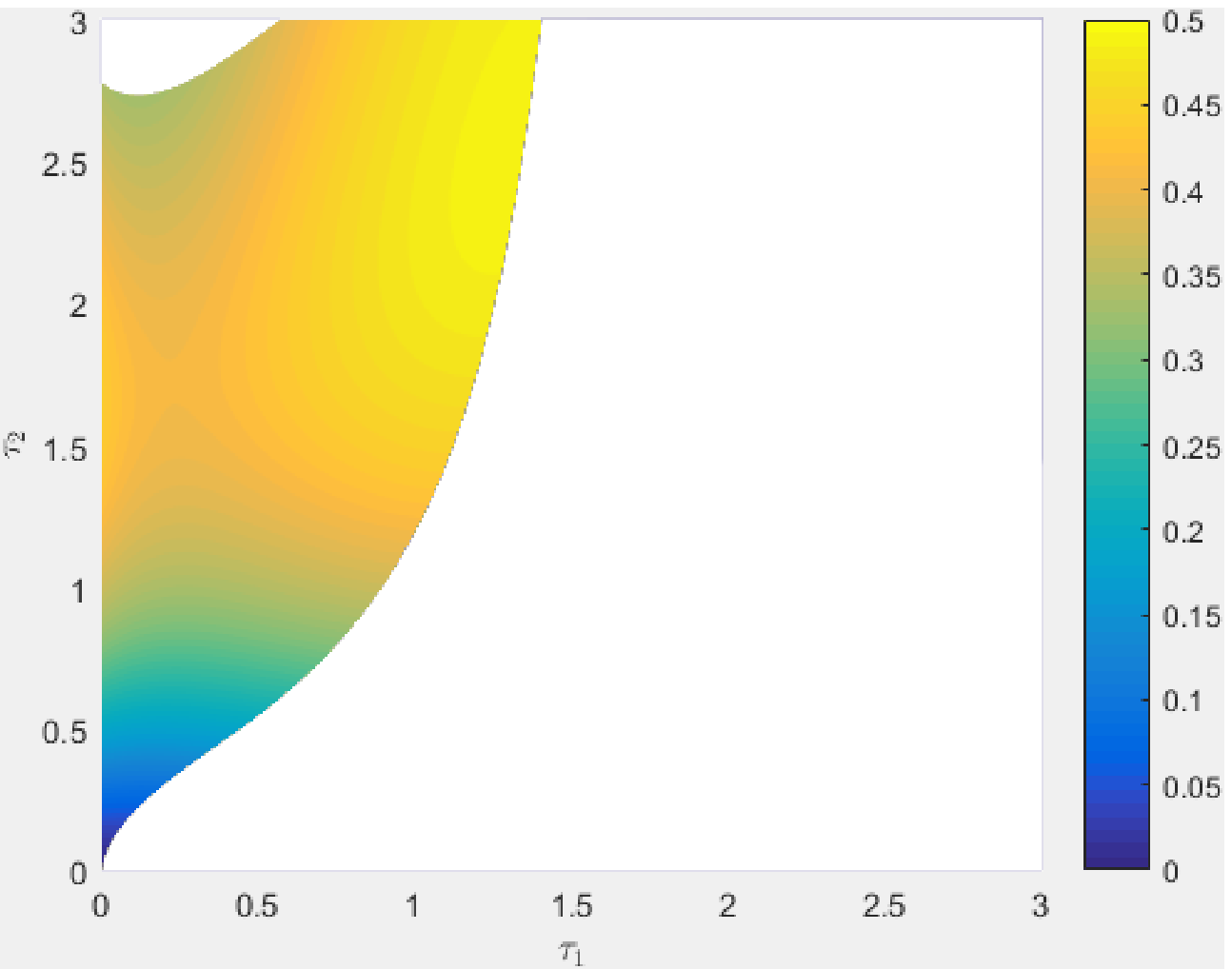}
        \caption{}
        \label{fig:ld16}
    \end{subfigure}
    \caption{influence of $l_{max}/d$ on feasible set, when
		(a) unconstrained; (b)  $l_{max}/d = 1$; (c)  $l_{max}/d = 2$; (d)  $l_{max}/d = 4$; (e)  $l_{max}/d = 8$; (f)  $l_{max}/d = 16$; 
		}
\label{ld fs}%
\end{figure}

By trying to minimize actuation during stepping, we find a unique sequence of step. Similar with our discussion in Section~\ref{vary kl}, this step sequence is whether a step time sequence or a step length sequence, and its classification is a result of different feasible sets~\eqref{eq:feasible set kmin}. This time, the feasible set is determined by $l_{max}/d$, as in Fig.~\ref{ld fs}. For a given robot, $l_{max}$ is fixed, so $d$ is the only factor. If $d$ is small enough, the feasible set will be large enough to include the global optimum of $f_{obj}$; in this case, it is associated with a step time sequence. If $d$ is large enough, the feasible set will become such small that the  global  optimum of $f_{obj}$ lives outside the feasible set; in this case, it is associated with a step length sequence, and each step is maximum step. An extremely large $d$ will result in empty feasible set, meaning the state is not $N$-step capturable. For a $N$-step case, we can find a decision boundary $d_{N,d}$ that classifies this step time/length sequence, if its state is $N$-step capturable. Furthermore, we are able to apply this $d_{N,d} \left( N=1, \cdots, n \right)$ in our algorithm for step planning. 

Given a robot with coefficient $k$ and $l_{max}$, we are able to find maximum disturbance for $N$-step $d_{N,max}$ based on our analysis in Section~\ref{vary kl}, and step time/length decision boundary for $N$-step $d_{N,d}$. Given a disturbance $d$, with a few comparisons between $d_{N,max}$ and $d_{N,d}$, we can plan step sequence for the robot. $d_{N,d}$ answers the question whether its state is $N$-step capturable, and $d_{N,d}$ answers whether it should follow a step time sequence or a step length sequence. The algorithm is a decision tree, and an example for 2-step planning is illustrated in Fig.~\ref{fig:flowchart}. Remind that whether we should look for more steps is dependent on $k$-$l_{max}$, and this flowchart corresponds to cases where 2-step plan is enough. In other cases where stepping more steps is helpful, the flowchart should expand.
\input{flowchart}

%% file: flowchart.tex
\begin{figure}%
\includegraphics[width=\columnwidth]{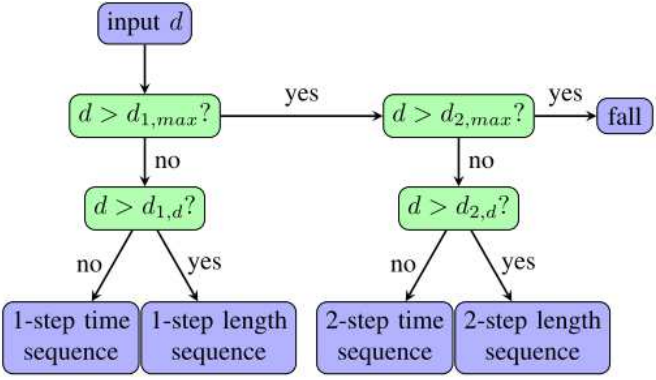}%
\caption{flowchart for 2-step push recovery algorithm\\
} \label{fig:flowchart}
\end{figure}

%% file: discussion.tex
\section{Discussion}

\subsection{capturability definition with time-margin}
We update original capturability definition to a novel notion of capturability with time-margin, and use this definition as framework for our analysis. Besides considering stance leg kinematics, time-margin capturable also takes in account swing leg kinematics, which concerns how fast swing leg moves.

Capturability in time-margin notion relates a step with its following steps. It ensures that a ($\Delta_N \cdots \Delta_1$)-margin $N$-step capturable state is ($\Delta_{N-1} \cdots \Delta_1$)-margin $(N-1)$-step capturable if the robot makes its first step under requirements of $\Delta_N$. This brings the time-varying capture region in~\cite{koolen2012capturability} to a static sequence of step positions.

Time-margin capturability offers more information than original definition. If a state is ($\Delta_N \cdots \Delta_1$)-margin $N$-step capturable, then there is a step sequence that contains $N$ steps and captures the robot. Time-margin definition also tells when to finish these $N$ steps by specifying $\Delta_N \cdots \Delta_1$, and further where to step from swing leg kernel.

\subsection{swing leg kernel}
The key insight of this paper is introducing swing leg kernel in analysis. Swing leg kernel describes swing leg kinematics, and is the bridge between step time and step length. It maps $N$-dimensional time domain to 2D Euclidean space where the robot lives. 

Different from previous works with simple assumptions on swing leg, we use a kernel that is a power function of step time. This swing leg kernel is mathematically convenient, and basically makes sense. With a model that combines a linear inverted pendulum model and a swing leg kernel, we are able to analyze influence of different swing leg kernels on capturability, and give suggestions on robot design regarding step ability. We also answer the question of how to step under a disturbance. Our mathematical approach is based on optimization. 

\subsection{enlightenment on robot design}
Based on our analysis on capturability with power function describing swing leg kernel, we give some suggestions that enlarge capturability of a robot. The maximum disturbance that a robot is able to resist acts as a measurement of capturability. 

We hold that actuation and maximum step length of a robot influence its capturability. In our model, actuation coefficient $k$ and normalized maximum step length $l_{max}$ represent these two factors. Due to variations in swing leg kernel, our conclusions are quantitative rather than qualitative. 

More powerful actuation gives larger capturability. In our context, we represent actuation as $k$. $k$ describes factors that relate to actuation, such as motor torque and leg moment of inertial. With fixed $l_{max}$, we find that capturability increases as $k$ increases. Using more powerful motors and reducing leg moment of inertial are two ways to increase $k$.

Longer normalized maximum step length $l_{max}$ gives larger capturability. So, generally, it is better to choose larger step length when designing a robot. For example, choose greater maximum angle between two legs. Moreover, as $l_{max}$ is $L_{max}$ normalized by height of center of mass, it  also benefits to lower center of mass. 


\subsection{step plan by minimizing actuation}
Given a robot, with $k_{max}$ and $l_{max}$, and a disturbance $d$, we can find a step time/length sequence that capture the robot in least steps with least actuation. If a state is $N$-step capturable, there exists whether a fixed step time sequence or a fixed step length time sequence that minimizes actuation.
This step time/length classification is dependent on $d$ value. If $d$ is large, it will step at maximum normalized step length at each step; if $d$ is small, it will follow the step time sequence corresponding to the global optimum of objective function. 

Inherent mathematics behind this classification is influence of $d$ on shape of feasible time domain. As step time and step length are related by swing leg kernel, our terminologies of `step time sequence' or `step length sequence' only represent mathematical coming of these step sequences. A small $d$ results in a large feasible time domain, and the global optimum of objective function is included in the feasible set. In this case, each step will take a period of time that corresponds to the global optimum. If $d$ is large, feasible set will become too small to include the global optimum, and stepping at maximum step length then becomes the best choice.

\subsection{whether two step plan is enough}
A recent heated discussion in robot step planning is whether two step plan is enough.  Based on an inverted pendulum model, Zaytsev et al. \cite{zaytsev2015two} concluded that two step is enough. They also supported their conclusion with evidence from other people, from both robotics  and biomechanics. 

Our answer to this question is: it depends. It depends on the swing leg kernel of the robot, but generally 2-step is enough. Among our selected swing leg kernels, maximum disturbance that a robot is able to resist generally  grows a little from 2-step to 3-step. The only exception is powerful actuation and short maximum step length. In this case, taking more steps still enlarges capturability. This point is also suggested in \cite{zaytsev2015two} by Art Kuo.

\subsection{longer step length to prevent falling}
We find our results fruitful in training elderly people to prevent falling. Falling is a major public health concern, especially for elderly people\cite{tinetti1988risk}. Mainstream training regarding this concerns muscle exercise, which is comparable to enlarging actuation coefficient in our swing leg kernel. 

In comparison, maximum step length fails to evoke enough attention. Our results also suggest larger step length\footnote{In our model, step width is equivalent to step length, as $x$ and $y$ axis are decoupled.} for less fall risks. Gabell and Nayak suggested that an increase in step width will lead to greater stability\cite{gabell1984effect}. Barak et al. found that fallers has smaller stride length, and show less stable gait patterns\cite{barak2006gait}. We give a model-based approach to support these results in gerontology, and we hold that exercise should be conducted both in kinesiology and in psychology to increase step length in order to reduce fall risks.

\section{Conclusion}
In this paper, we expand original definition of capturability to a notion that is associated with a time-margin. We further introduce a swing leg kernel to describe swing leg kinematics. We analyze $N$-step capturability with a combination of swing leg kernel and a linear inverted pendulum model.  We conclude that more powerful actuation and larger step length result in greater capturability. We also find a step time or a step length sequence classification, based on boundary value problem analysis.

%% file: appendix.tex
\appendices

\section{Proof for 1-step Capture Point}
\label{proof uniqueness}

For LIPM with point foot, the instantaneous capture point is the unique 1-step capture point, if it is able to step to any location at any time.
\begin{IEEEproof}
First show that the instantaneous capture point is a 1-step capture point. 

If a step is made at $r_{ic}$ ($i.e. \  r_{ankle} := r_{ic}$), right hand side of Equation~\eqref{eq: dyn 3}  is zero. This gives $\dot{r}_{ic}$ to be zero. we also have
\begin{eqnarray}
\ddot{r}_{CoM}&= & \mathbf{P}r_{CoM} - r_{ic} \nonumber \\ 
&= & \mathbf{P}r_{CoM} - \left(\mathbf{P}r_{CoM}  + \dot{r}_{CoM}\right) \nonumber \\ 
&= & -\dot{r}_{CoM}.
\label{eq:r_com dyn}
\end{eqnarray}
The system given by Equation~\eqref{eq:r_com dyn} has eigenvalues -1, so $\dot{r}_{CoM}$ vanishes. Thus by Equation~\eqref{eq:dyn 1}, $\ddot{r}_{CoM} \rightarrow 0$, which further implies $\mathbf{P}r_{CoM} \rightarrow r_{ankle}$. As $r_{CoM}$, $r_{ankle}$ and $r_{ic}$ converge, the robot is captured.

Second, we use Chetaev instability theorem to show the 1-step capture point is unique. Assume that there is another 1-step capturable point, $i.e.$ $r_{ankle}' \neq r_{ic}$.  Whenever a step is finished, $r_{ankle}'$ is constant. Thus $\dot{r}_{ankle}'$ is 0. Taking time derivative on both sides of Equation~\eqref{eq: dyn 3}, we have
\begin{eqnarray}
\ddot{r}_{ic} &=& \dot{r}_{ic} \nonumber \\
&=& r_{ic} - r_{ankle}'
\label{eq:}
\end{eqnarray}
Choose a Lyapunov fuction as 
\begin{equation}
V = \frac{1}{2}\dot{r}_{ic}^2
\label{eq:}
\end{equation}
$V>0$, if the $\dot{r}_{ic}$ is nonzero, and $V(0) = 0$. $V$  is radically unbounded. Its derivative is 
\begin{eqnarray}
\dot{V} &=& \dot{r}_{ic}\ddot{r}_{ic} \nonumber \\
&=& (r_{ic} - r_{ankle}')^2 
\label{eq:}
\end{eqnarray}
If $r_{ankle}' \neq r_{ic}$, $\dot{V} > 0$. By Chetaev instability theorem, the system is unstable. This implies that $r_{ankle}'$ is not a 1-step capture point.

With above, we conclude that the instantaneous capture point is the unique 1-step capture point for LIPM with point foot.
\end{IEEEproof}

\section{Swing Leg Kernel from Simulation}
\label{appendix:slk}

\begin{figure}%
\centering
\def\svgwidth{0.5\columnwidth}
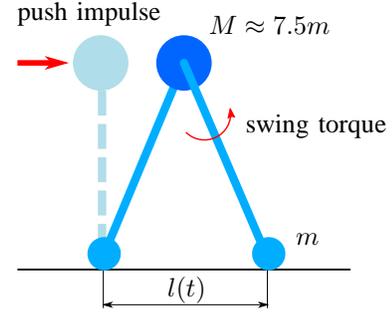
\caption{simulation model to determine swing leg kernel}%
\label{fig:appendix2}%
\end{figure}

\begin{figure}%
\centering
\includegraphics[width=0.75\columnwidth]{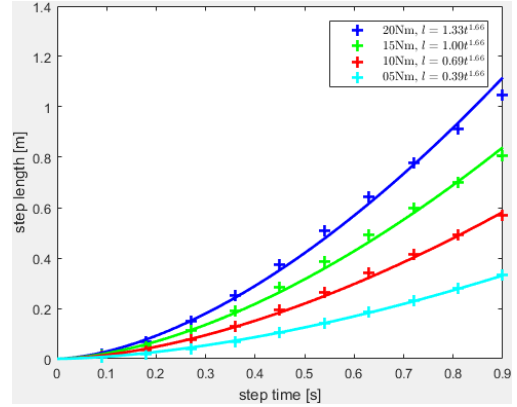}%
\caption{power function approximation of different swing torques}%
\label{30impulse}%
\end{figure}

Our simulation model is depicted in Fig.~\ref{fig:appendix2}. Simulation results indicate that the step length is a power function of step time approximately.
We use a constant torque to swing the leg, and record relation between $\tau$ and $l$. We then find that a power function is able to approximate the data. We also simulate for a various value of swing torques. For different swing torques, we find that index number $a$ is always around 1.66, and $k$ increases as swing torque increases. Fig.~\ref{30impulse} shows approximation for an impulse of  30 N at 0.1 s.

%% file: drawing2.eps_tex
\begingroup%
  \makeatletter%
  \providecommand\color[2][]{%
    \errmessage{(Inkscape) Color is used for the text in Inkscape, but the package 'color.sty' is not loaded}%
    \renewcommand\color[2][]{}%
  }%
  \providecommand\transparent[1]{%
    \errmessage{(Inkscape) Transparency is used (non-zero) for the text in Inkscape, but the package 'transparent.sty' is not loaded}%
    \renewcommand\transparent[1]{}%
  }%
  \providecommand\rotatebox[2]{#2}%
  \ifx\svgwidth\undefined%
    \setlength{\unitlength}{480bp}%
    \ifx\svgscale\undefined%
      \relax%
    \else%
      \setlength{\unitlength}{\unitlength * \real{\svgscale}}%
    \fi%
  \else%
    \setlength{\unitlength}{\svgwidth}%
  \fi%
  \global\let\svgwidth\undefined%
  \global\let\svgscale\undefined%
  \makeatother%
  \begin{picture}(1,0.93466917)%
    \put(0,0){\includegraphics[width=\unitlength]{drawing2.eps}}%
    \put(0,0.85){\color[rgb]{0,0,0}\makebox(0,0)[lb]{push  impulse}}%
    \put(0.68571157,0.53){\color[rgb]{0,0,0}\makebox(0,0)[lb]{\smash{swing torque}}}%
    \put(0.45,0.03745079){\color[rgb]{0,0,0}\makebox(0,0)[lb]{\smash{$l(t)$}}}%
    \put(0.83607244,0.19242994){\color[rgb]{0,0,0}\makebox(0,0)[lb]{\smash{$m$}}}%
    \put(0.57505473,0.82662153){\color[rgb]{0,0,0}\makebox(0,0)[lb]{\smash{$M \approx 7.5m$}}}%
  \end{picture}%
\endgroup%

%% file: capturability_notes.bbl
\begin{thebibliography}{10}
\providecommand{\url}[1]{#1}
\csname url@samestyle\endcsname
\providecommand{\newblock}{\relax}
\providecommand{\bibinfo}[2]{#2}
\providecommand{\BIBentrySTDinterwordspacing}{\spaceskip=0pt\relax}
\providecommand{\BIBentryALTinterwordstretchfactor}{4}
\providecommand{\BIBentryALTinterwordspacing}{\spaceskip=\fontdimen2\font plus
\BIBentryALTinterwordstretchfactor\fontdimen3\font minus
  \fontdimen4\font\relax}
\providecommand{\BIBforeignlanguage}[2]{{%
\expandafter\ifx\csname l@#1\endcsname\relax
\typeout{** WARNING: IEEEtran.bst: No hyphenation pattern has been}%
\typeout{** loaded for the language `#1'. Using the pattern for}%
\typeout{** the default language instead.}%
\else
\language=\csname l@#1\endcsname
\fi
#2}}
\providecommand{\BIBdecl}{\relax}
\BIBdecl

\bibitem{vukobratovic1969contribution}
M.~Vukobratovic and D.~Juricic, ``Contribution to the synthesis of biped
  gait,'' \emph{Biomedical Engineering, IEEE Transactions on}, no.~1, pp. 1--6,
  1969.

\bibitem{vukobratovic2004zero}
M.~Vukobratovi{\'c} and B.~Borovac, ``Zero-moment point—thirty five years of
  its life,'' \emph{International Journal of Humanoid Robotics}, vol.~1,
  no.~01, pp. 157--173, 2004.

\bibitem{htirmiizlii1987bipedal}
Y.~Htirmiizlii and G.~D. Moskowitz, ``Bipedal locomotion stabilized by impact
  and switching: I. two-and three-dimensional, three-element models,''
  \emph{Dynamics and Stability of Systems}, vol.~2, no.~2, 1987.

\bibitem{mcgeer1990passive}
T.~McGeer, ``Passive dynamic walking,'' \emph{the international journal of
  robotics research}, vol.~9, no.~2, pp. 62--82, 1990.

\bibitem{goswami1996limit}
A.~Goswami, B.~Espiau, and A.~Keramane, ``Limit cycles and their stability in a
  passive bipedal gait,'' in \emph{Robotics and Automation, 1996. Proceedings.,
  1996 IEEE International Conference on}, vol.~1.\hskip 1em plus 0.5em minus
  0.4em\relax IEEE, 1996, pp. 246--251.

\bibitem{tedrake2004actuating}
R.~Tedrake, T.~W. Zhang, M.-f. Fong, and H.~S. Seung, ``Actuating a simple 3d
  passive dynamic walker,'' in \emph{Robotics and Automation, 2004.
  Proceedings. ICRA'04. 2004 IEEE International Conference on}, vol.~5.\hskip
  1em plus 0.5em minus 0.4em\relax IEEE, 2004, pp. 4656--4661.

\bibitem{grizzle2001asymptotically}
J.~W. Grizzle, G.~Abba, and F.~Plestan, ``Asymptotically stable walking for
  biped robots: Analysis via systems with impulse effects,'' \emph{Automatic
  Control, IEEE Transactions on}, vol.~46, no.~1, pp. 51--64, 2001.

\bibitem{koolen2012capturability}
T.~Koolen, T.~De~Boer, J.~Rebula, A.~Goswami, and J.~Pratt,
  ``Capturability-based analysis and control of legged locomotion, part 1:
  Theory and application to three simple gait models,'' \emph{The International
  Journal of Robotics Research}, vol.~31, no.~9, pp. 1094--1113, 2012.

\bibitem{pratt2012capturability}
J.~Pratt, T.~Koolen, T.~De~Boer, J.~Rebula, S.~Cotton, J.~Carff, M.~Johnson,
  and P.~Neuhaus, ``Capturability-based analysis and control of legged
  locomotion, part 2: Application to m2v2, a lower body humanoid,'' \emph{The
  International Journal of Robotics Research}, p. 0278364912452762, 2012.

\bibitem{zaytsev2015two}
P.~Zaytsev, S.~J. Hasaneini, and A.~Ruina, ``Two steps is enough: no need to
  plan far ahead for walking balance,'' in \emph{Robotics and Automation, 2015.
  Proceedings., 2015 IEEE International Conference on}.\hskip 1em plus 0.5em
  minus 0.4em\relax IEEE, 2015.

\bibitem{tinetti1988risk}
M.~E. Tinetti, M.~Speechley, and S.~F. Ginter, ``Risk factors for falls among
  elderly persons living in the community,'' \emph{New England journal of
  medicine}, vol. 319, no.~26, pp. 1701--1707, 1988.

\bibitem{gabell1984effect}
A.~Gabell and U.~Nayak, ``The effect of age on variability in gait,''
  \emph{Journal of Gerontology}, vol.~39, no.~6, pp. 662--666, 1984.

\bibitem{barak2006gait}
Y.~Barak, R.~C. Wagenaar, and K.~G. Holt, ``Gait characteristics of elderly
  people with a history of falls: a dynamic approach,'' \emph{Physical
  therapy}, vol.~86, no.~11, pp. 1501--1510, 2006.

\end{thebibliography}
